\documentclass[review]{elsarticle}

\usepackage{lineno,hyperref}

\usepackage{framed,multirow}

\usepackage{amssymb}
\usepackage{latexsym}

\usepackage{url}
\usepackage{xcolor}

\usepackage{hyperref}
\usepackage{amsmath,amssymb,amsfonts}
\usepackage{graphicx}
\usepackage{textcomp}
\usepackage{multirow}
\usepackage{caption}
\usepackage{url}
\usepackage{hyperref}
\usepackage{soul}
\usepackage{cleveref}
\usepackage{makecell}
\usepackage{xcolor}
\usepackage{tikz}
\usepackage{float}
\usepackage{soul}

\usepackage{latexsym}
\usepackage{rotating}
\usepackage{url}
\usepackage{xcolor}
\usepackage{amsmath}
\usepackage{makecell}

\usepackage{hyperref}
\usepackage{soul}
\usepackage{cleveref}
\usepackage{enumitem}
\usepackage{algorithm}

\usepackage{algpseudocode}
\modulolinenumbers[1]

\journal{Journal of Medical Image Analysis}

\definecolor{newcolor}{rgb}{.8,.349,.1}
\begin{document}

%\verso{Wu \textit{et~al.}}

\begin{frontmatter}

\title{GAMMA Challenge:
Glaucoma grAding from Multi-Modality imAges}%
% \tnotetext[tnote1]{This is an example for title footnote coding.}

\author[2]{Junde Wu\fnref{my_note}\corref{cor1}}
\author[2]{Huihui Fang\fnref{my_note}\corref{cor1}}
%e-mail:fanghuihui@baidu.com
\author[1]{Fei Li\fnref{my_note}\corref{cor1}}
%e-mail:lifei57@mail.sysu.edu.cn
\author[3]{Huazhu Fu\fnref{my_note}}
%e-mail:hzfu@ieee.org

\author[1]{Fengbin Lin}
%e-mail:woalsdnd@gmail.com
\author[5]{Jiongcheng Li}
%e-mail:shirlyyu@tencent.com
\author[5]{Yue Huang}
%yhuang2010@xmu.edu.cn
%e-mail:zhangmenglu2018@email.szu.edu.cn
\author[6]{Qinji Yu}
%e-mail:923241961@qq.com
\author[7]{Sifan Song}
%e-mail:tronbian@tencent.com
\author[8]{Xinxing Xu}
\author[8]{Yanyu Xu}
%e-mail:leiby@szu.edu.cn
\author[9]{Wensai Wang}
%e-mail:1810272021@email.szu.edu.cn
\author[9]{Lingxiao Wang}
%e-mail:xuxinx@ihpc.a-star.edu.sg
\author[10]{Shuai Lu}
%e-mail:li_shaohua@ihpc.a-star.edu.sg
\author[10,19]{Huiqi Li}
%e-mail:ffumerob@ull.edu.es
\author[11]{Shihua Huang}
%e-mail:jfsigut@ull.edu.es 
\author[12]{Zhichao Lu}
%e-mail:h.almubarak@ieee.org
\author[13]{Chubin Ou}
%e-mail:ybazi@ksu.edu.sa
\author[13]{Xifei Wei}
%e-mail:yuanhao.guo@ia.ac.cn
\author[14]{Bingyuan Liu}
%e-mail:zhouyating2020@ia.ac.cn
\author[15]{Riadh Kobbi}
%e-mail:ujjwalbaid0408@gmail.com
\author[16]{Xiaoying Tang}
%e-mail:shubham.innani@gmail.com 
\author[16,17]{Li Lin}
%e-mail:292725651@sjtu.edu.cn
\author[18]{Qiang Zhou}
%e-mail:jieyang@sjtu.edu.cn
\author[18]{Qiang Hu}
%e-mail:jieyang@sjtu.edu.cn
\author[20]{Hrvoje Bogunovi\'c}\fnref{my_note}
%e-mail:jieyang@sjtu.edu.cn
\author[21]{José~Ignacio~Orlando}\fnref{my_note} 
%e-mail:jieyang@sjtu.edu.cn
\author[1]{Xiulan Zhang}\fnref{my_note}\corref{cor2} 
%e-mail:zhangxl2@mail.sysu.edu.cn
\author[2]{Yanwu Xu}\fnref{my_note}\corref{cor2}  
%e-mail:xuyanwu@baidu.com

\address[1]{State Key Laboratory of Ophthalmology, Zhongshan Ophthalmic Center, Sun Yat-sen University, Guangdong Provincial Key Laboratory of Ophthalmology and Visual Science, Guangzhou, China}
\address[2]{Intelligent Healthcare Unit, Baidu Inc., Beijing, China}
\address[3]{Institute of High Performance Computing (IHPC), Agency for Science, Technology and Research (A*STAR), Singapore}
% \address[4]{VUNO Inc.}
\address[5]{School of Informatics, Xiamen University, Xiamen, China}
\address[6]{Shanghai Jiao Tong University, Shanghai, China}
\address[7]{Xi'an Jiaotong-Liverpool University, Suzhou, China}
\address[8]{Institute of High Performance Computing,A*STAR, Singapore}
\address[9]{Institute of Biomedical Engineering, Chinese Academy of Medical Sciences and Peking Union Medical College, Tianjin, China}
\address[10]{ School of Medical Technology, Beijing Institute of Technology, Beijing, China}
\address[11]{Department of Computing, Hong Kong Polytechnic University, Hong Kong, China}
\address[12]{Department of Computer Science and Engineering, Southern University of Science and Technology, Shenzhen, China}
\address[13]{Weizhi Medical Technology Company, Suzhou, China}
\address[14]{École de technologie supérieure, Montreal, Montreal, Canada}
\address[15]{DIAGNOS Inc., Quebec, Canada}
\address[16]{Department of Electrical and Electronic Engineering, Southern University of Science and Technology, Shenzhen, China}
\address[17]{Department of Electrical and Electronic Engineering, The University of Hong Kong, Hong Kong, China}
\address[18]{Suixin (Shanghai) Technology Co., Ltd., Shanghai, China}
\address[19]{School of Information and Electronics, Beijing Institute of Technology, Beijing, China}
\address[20]{Christian Doppler Lab for Artificial Intelligence in Retina, Department of Ophthalmology, Medical University of Vienna, Austria}
\address[21]{Yatiris Group, PLADEMA Institute, CONICET, UNICEN, Tandil, Argentina}

\cortext[cor1]{These authors contributed equally to the work.} 
\cortext[cor2]{Corresponding authors: Xiulan Zhang (zhangxl2@mail.sysu.edu.cn), and Yanwu Xu (xuyanwu@baidu.com).}
\fntext[my_note]{These authors co-organized the GAMMA challenge. All others contributed results of their algorithms presented in the paper.}

% \received{***}
% \finalform{***}
% \accepted{***}
% \availableonline{***}

\begin{abstract}
Glaucoma is a chronic neuro-degenerative condition that is one of the world's leading causes of irreversible but preventable blindness. The blindness is generally caused by the lack of timely detection and treatment. Early screening is thus essential for early treatment to preserve vision and maintain life quality. Color fundus photography and Optical Coherence Tomography (OCT) are the two most cost-effective tools for glaucoma screening. Both imaging modalities have prominent biomarkers to indicate glaucoma suspects, such as the vertical cup-to-disc ratio (vCDR) on fundus images and retinal nerve fiber layer (RNFL) thickness on OCT volume. In clinical practice, it is often recommended to take both of the screenings for a more accurate and reliable diagnosis. However, although numerous algorithms are proposed based on fundus images or OCT volumes for the automated glaucoma detection, there are few methods that leverage both of the modalities to achieve the target.
% Inspired by the success of Retinal Fundus Glaucoma Challenge (REFUGE) (\cite{orlando2020refuge})\HBc{I would not say this in the abstract} we held previously,
To fulfill the research gap, we set up the Glaucoma grAding from Multi-Modality imAges (GAMMA) Challenge to encourage the development of fundus \& OCT-based glaucoma grading. The primary task of the challenge is to grade glaucoma from both the 2D fundus images and 3D OCT scanning volumes. As part of GAMMA, we have publicly released a glaucoma annotated dataset with both 2D fundus color photography and 3D OCT volumes, which is the first multi-modality dataset for machine learning based glaucoma grading. In addition, an evaluation framework is also established to evaluate the performance of the submitted methods. During the challenge, 1272 results were submitted, and finally, ten best performing teams were selected for the final stage. We analyze their results and summarize their methods in the paper. Since all the teams submitted their source code in the challenge, we conducted a detailed ablation study to verify the effectiveness of the particular modules proposed. Finally, we identify the proposed techniques and strategies that could be of practical value for the clinical diagnosis of glaucoma. As the first in-depth study of fundus \& OCT multi-modality glaucoma grading, we believe the GAMMA Challenge will serve as an essential guideline and benchmark for future research.
\end{abstract}

% \begin{keyword} 
% \KWD Glaucoma \sep Color fundus photography\sep Multi-modality \sep Optical coherence tomography \sep GAMMA Challenge
% \end{keyword}

\end{frontmatter}

%\linenumbers

%% main text
\section{Introduction}
\label{sec:introduction}

\begin{figure}[ht]
\centering
\includegraphics[width=1\linewidth]{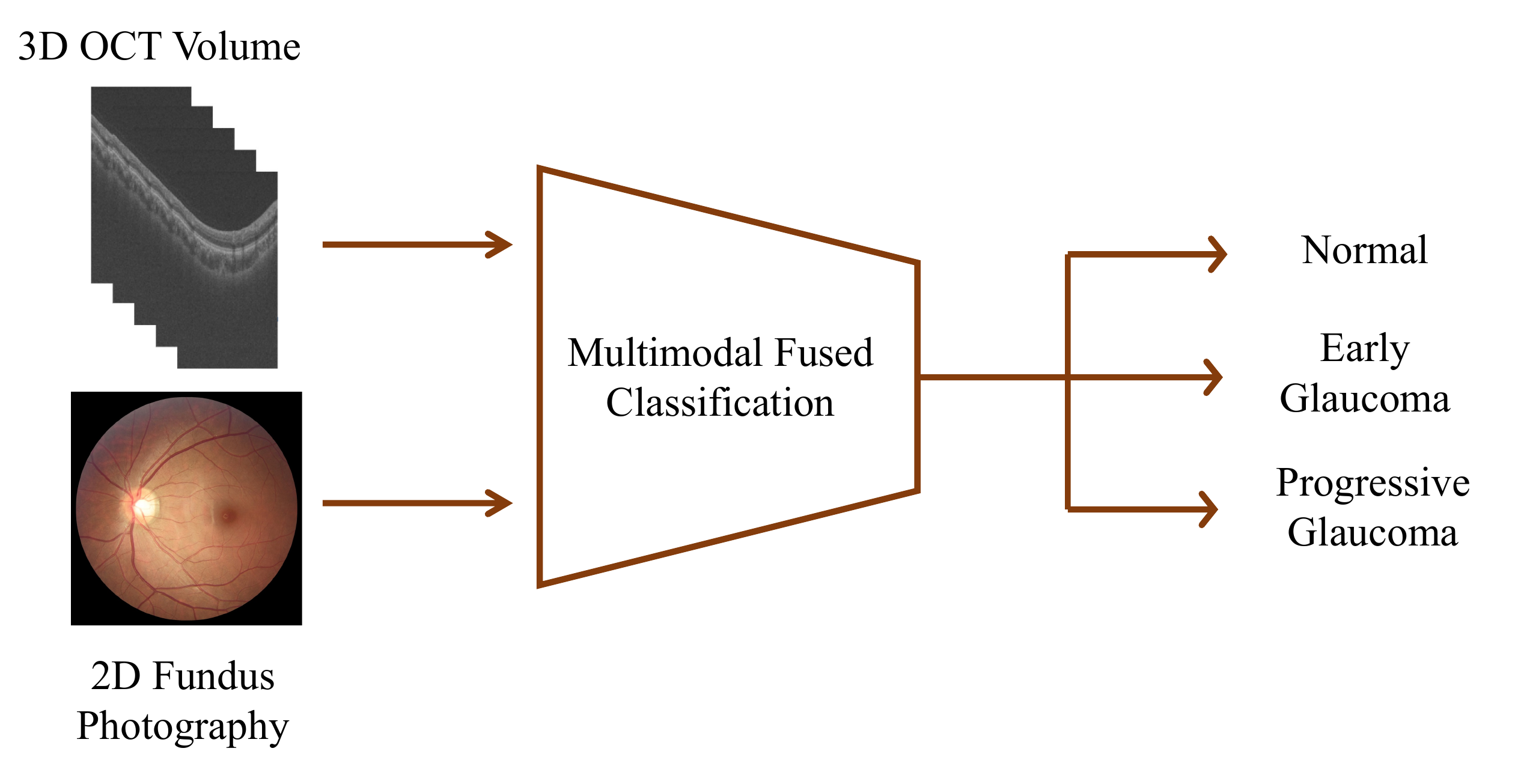}
\caption{An illustration of the GAMMA Challenge. The primary goal of the challenge is to predict the cases as normal, early-glaucoma or progressive-glaucoma from fundus-OCT pairs.}
\label{fig:gamma}
\end{figure}

Worldwide, glaucoma is the second-leading cause of blindness after cataracts (\cite{resnikoff2004global}). About 70 million people have glaucoma globally (\cite{vos2016global}). Glaucoma can occur without any cause, but is affected by many factors. The most important of which is the intra-ocular eye pressure (IOP). Aqueous humor in the eyes flows through the pupil to the front of the eye. In a healthy eye, the fluid leaves through a drainage canal located between the iris and cornea. With glaucoma, the drainage canals become clogged with microscopic deposits. The fluid builds up in the eye. This excess fluid puts pressure on the eye. Eventually, this elevated eye pressure can damage the optic nerve head (ONH) leading to glaucoma.

Many forms of glaucoma have no warning signs. The effect is so gradual that one may not notice a change in vision until the condition is at an advanced stage. That is why glaucoma is also called the 'silent thief of sight'. Because vision loss due to glaucoma can not be recovered, it is important for the early diagnosis. If glaucoma is recognized early, vision loss can be slowed or prevented.

% \HBc{I think a longer intro is needed on what glaucoma is, how it manifests itself and why screering and early diagnosis is so important}. 
The function-based visual field test is the clinical gold standard of glaucoma screening, but it does not show signs of early glaucoma. Instead, an optic nerve head (ONH) assessment is a convenient way to detect early glaucoma and is currently performed widely for glaucoma screening (\cite{jonas1999ophthalmoscopic,morgan2005digital,fu2017segmentation}). As practical and noninvasive tools, 2D fundus photography and 3D optical coherence tomography (OCT) are the most commonly used imaging modalities to evaluate the optic nerve structure in clinical practice. 
%However, manual ONH assessment by trained clinicians is time-consuming, costly, and objective. There is an increasing need for an objective evaluation of the optic nerve structure, particularly for preperimetric glaucoma. 

The main advantage of the fundus photographs is that they can clearly show the optic disc, optic cup, and blood vessels. Among them, the clinical parameters like the vertical cup to disc ratio (vCDR), disc diameter, and the ratio of blood vessels area in inferior-superior side to area of blood vessel in the nasal-temporal side have been validated to be of great significance for the glaucoma diagnosis (\cite{jonas2000ranking,hancox1999optic,nayak2009automated,li2022dynamic}). OCT measures retinal nerve fiber layer (RNFL) thickness based on its optical properties. RNFL thickness, computed from OCT volumes that are acquired in cylindrical sections surrounding the optic disc, is often used to identify glaucoma suspect. Though OCT volumes and fundus photographs are effective tools for diagnosing early glaucoma, neither of them alone can be used to exclude it. Clinically, ophthalmologists often recommend to take both of the screenings for a more accurate and reliable diagnosis. Recent report shows nearly 46.3\% glaucoma cases would be ignored if using fundus images or OCT volume alone (\cite{anton2021diagnostic}).

However, in terms of computer-aided glaucoma diagnosis, most algorithms are developed on only single modality. Although fundus photographs and OCT are both the mainstream glaucoma screening tools in clinical practice, few algorithms are established that make use of both modalities. This is primarily due to two reasons: a) there is no publicly available dataset to train and evaluate such models, and b) due to the discrepancy in the characteristics and the dimensionality between the two modalities, the task is technically challenging. 
%In addition, although screening the early-stage glaucoma is urgently needed clinically, most existing algorithms focused only on the binary classification (normal/glaucoma) in glaucoma diagnosis.

In order to overcome these issues, a challenge with an dataset , as a way to encourage the development of SOTA imaging technology on this clinically relevant task, may be an appropriate approach. Inspired by the success of Retinal Fundus Glaucoma Challenge (REFUGE) (\cite{orlando2020refuge}) we previously held, the Glaucoma grAding from Multi-Modality imAges (GAMMA) Challenge was organized in conjunction with the 8th Ophthalmic Medical Image Analysis (OMIA) workshop, during MICCAI 2021 (Strasbourg, France) to encourage the development of fundus \& OCT-based multi-modal glaucoma grading algorithms.
%Participants are encouraged to design and propose state-of-the-art methods for fundus images and OCT based multi-modality glaucoma grading. 
Given a pair consisting of a fundus image and an OCT volume, the submitted algorithms need to predict the case as normal, early-glaucoma, or progressive-glaucoma (intermediate and advanced stage). An illustration is shown in Figure \ref{fig:gamma}. We also describe an evaluation framework to rank the participated teams. Ten top performing teams were invited to share their technical reports and source code. In brief, the primary contribution of the GAMMA Challenge is two-fold: 

a) The first publicly available multi-modality glaucoma grading dataset for deep learning based methods is released, providing fundus photography and OCT volume pairs. 

b) State-of-the-art (SOTA) machine learning methods are evaluated to encourage the development of novel methodologies for fundus \& OCT-based glaucoma grading. \\
Due to the success of the challenge, GAMMA is expected to serve as the main benchmark for this clinically relevant task in the future.

\begin{figure}[ht]
\centering
\includegraphics[width=1\linewidth]{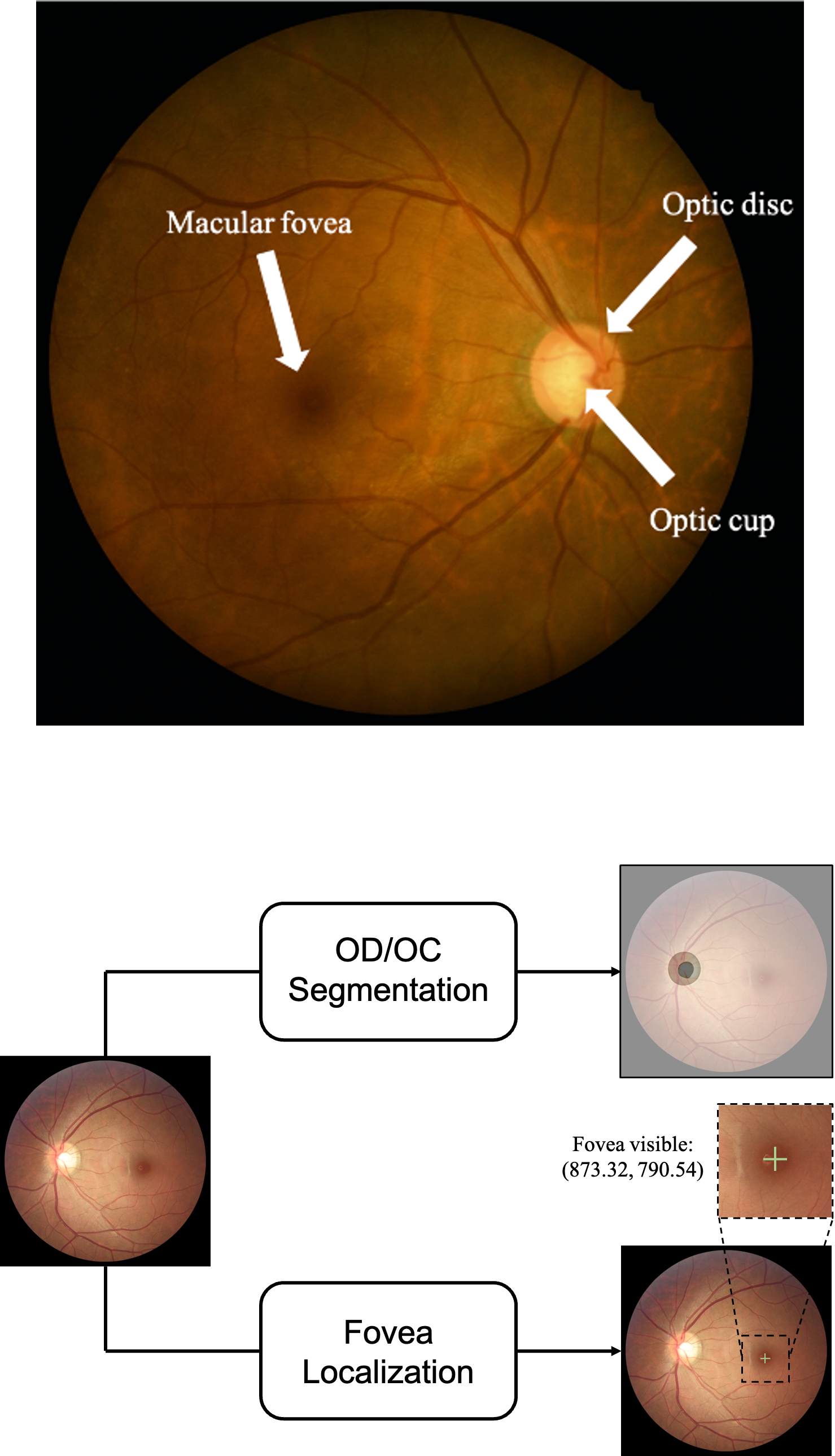}
\caption{An illustration of the GAMMA auxiliary tasks: optic disc/cup (OD/OC) segmentation and fovea localization on fundus images.}
\label{fig:aux}
\end{figure}

Besides glaucoma grading labels, the optic disc \& cup (OD/OC) mask labels as well as fovea location labels are also provided in the GAMMA dataset. These auxiliary tasks were proposed to investigate the role of optic disc and fovea in glaucoma grading. Thus, the participants can also submit algorithms for the OD/OC segmentation task and fovea localization task, and the final team performance includes the achieved scores on these auxiliary tasks. An illustration of the auxiliary tasks is shown in Figure \ref{fig:aux}. In the GAMMA Challenge, the participants are encouraged to utilize the auxiliary tasks to improve the performance of glaucoma grading. 

The inception of the GAMMA challenge encourages many participants to contribute SOTA machine learning techniques on this task. This manuscript summarizes the GAMMA Challenge, analyzes their results, and investigates their particular approaches. All top-10 teams submitted the source code of their algorithms. This allowed us to conduct a detailed ablation study to identify which techniques were the most effective ones for the screening task. We believe that our analysis of SOTA machine learning methods will greatly benefit the future algorithm design on this task.

\section{The GAMMA challenge}
\label{sect:adam}
%The GAMMA challenge was held in conjunction with the 8th Ophthalmic Medical Image Analysis (OMIA) workshop, during MICCAI 2021 (Strasbourg, France). 
The GAMMA challenge was officially launched from 20 Mar 2021 to 01 October 2021, which focuses on the field of glaucoma grading based on multi-modality images (Fundus photography and OCT volume). 
% We released a dataset containing 300 multi-modal acquisitions consisting of a fundus image and a 3D OCT volume. Moreover, we also provided a uniform evaluation framework for a fair comparison of different models. In addition to glaucoma grading, there are also two auxiliary tasks: optic disc/cup segmentation and fovea localization on the fundus images. An illustration of the auxiliary tasks is shown in Figure \ref{fig:aux}. Participants were encouraged to leverage the OD/OC and fovea information to improve the glaucoma grading performance.
The challenge consisted of a preliminary stage and a final stage. During the preliminary stage, we released a training set for the participating teams to train the models. The registered teams were allowed to use the training set to learn their proposed algorithms for glaucoma grading, and, optionally, for OD/OC segmentation and fovea localization. Their results can be submitted on \url{https://aistudio.baidu.com/aistudio/competition/detail/90/0/submit-result} and would be evaluated on the preliminary set. The registered teams then can see their performance on the preliminary set and adjust their algorithms. For a fair comparison of the proposed methods, the registered teams were not allowed to use any other private data set for developing their methods.

This preliminary stage lasted 30 days, and each team was allowed to make a maximum of five submissions per day. A total of 70 teams submitted 1272 valid results to the challenge platform during the preliminary stage, out of which ten teams, based on their method performance and the willingness to participate in the OMIA8 workshop, were selected to the final stage. The ten such selected teams were then ranked based on their performance on the final test set. For the final stage, teams were not allowed to modify their models anymore. 

% \vspace{-0.2cm}
\subsection{GAMMA Dataset}
\label{subsect:adamdata}
The dataset released by GAMMA was provided by Sun Yat-sen Ophthalmic Center, Sun Yat-sen University, Guangzhou, China, and the glaucoma and non-glaucoma subjects were randomly selected from glaucoma and myopia cohort, respectively. The dataset contained 300 samples of fundus-OCT pairs. %These two modalities are commonly used in the clinical fundus examination. 
The image acquisitions were performed in a standardized darkroom, and the patients were requested to sit upright. The OCT volumes were all acquired with a Topcon DRI OCT Triton. The OCT was centered on the macula, had a 3 $\times$ 3 mm en-face field of view, and each volume contained 256 two-dimensional cross-sectional images with a size of $992 \times 512$ pixels. The fundus images were acquired using a KOWA camera with a resolution of $2000 \times 2992$ pixels and a Topcon TRC-NW400 camera with a resolution of $1934 \times 1956$ pixels. The fundus images in our dataset were centered on the macula or on the midpoint between optic disc and macula, with both optic disc and macula visible. The image quality was checked manually. The 300 samples in the GAMMA dataset correspond to 276 Chinese patients (42\% female), which ranged in age from 19-77 and averaged at 40.64$\pm$14.53 years old. Glaucoma accounted for 50\% of the sample, including 52\% in the early stage, 28.67\% in the intermediate stage, and 19.33\% in the advanced stage. Early glaucoma samples were obtained from 64 patients with average age of 43.47$\pm$15.49, of whom 14 patients provided data from both eyes, another 30 patients provided data from the oculus sinister (OS), and 20 patients provided data from the oculus dexter (OD). Similarly, intermediate and advanced glaucoma samples were obtained from 35 and 27 patients with average ages of 47.98$\pm$17.38 and 46.24$\pm$14.47, respectively. In the intermediate glaucoma samples, 8 patients provided data from both eyes, 15 patients provided OS data, and 12 patients provided OD data. In the advanced glaucoma samples, 2 patients provided data from both eyes, 8 patients provided OS data, and 17 patients provided OD data. The non-glaucomatous samples in the dataset were collected from 150 patients with average age of 35.97$\pm$11.29, 57 and 93 patients provided OS and OD data, respectively.
We randomly divided the collected samples of each category (non-glaucoma, early-glaucoma, intermediate-glaucoma, and advanced-glaucoma) into three roughly equal parts and assigned them to each of the three challenge sets, corresponding to, we prepared 100 data pairs for each the training, preliminary process and final processes. Because the data sizes in intermediate and advanced glaucoma categories are relatively small compared to that of early glaucoma category, 
%providing data with category imbalance in model training is not conducive to model learning, 
so we grouped the intermediate and advanced glaucoma into one category, i.e., progressive-glaucoma in the main challenge tasks.
%The intermediate and advanced glaucoma were combined into a progressive-glaucoma class~\HBc{Can you add an explanation why this was combined?} for the main challenge task.
%We divided the collected samples across the three challenge sets equally\HBc{How was this done? Randomly?}, 

In addition, the GAMMA dataset included the respective glaucoma grades, the fovea coordinates, and the mask of the cup and optic disc. The GAMMA dataset is publicly available through \url{https://gamma.grand-challenge.org/}, and is allowed to be used and distributed under CC BY-NC-ND (Attribution-NonCommercial-NoDerivs) licence. The following sections describe the implementation of the annotation processes of the three challenge tasks.

\subsubsection{Glaucoma Grading}
The ground truth of glaucoma grading task for each sample was determined based on mean deviation (MD) values from visual field reports following the criteria below: early-stage with MD value higher than -6 dB, intermediate stage with MD value between -6 and -12 dB, advanced stage with MD value worse than -12 dB. These visual field reports were generated on the same day as the OCT examination and were reliable with fixation losses of under 2/13 and false-positive rate under 15\% and false-negative rate under 25\% (\cite{li2020development, xiong2021multimodal}). 

\subsubsection{Fovea Localization}\label{sec:foveaeval}
The initial fovea coordinate annotation of each fundus image was performed manually by four clinical ophthalmologists from Sun Yat-sen Ophthalmic Center, Sun Yat-sen University, China, who had an average of 8 years of experience in the field (range 5-10 years). All ophthalmologists independently located the fovea in the image using a cross marker without having access to any patient information or knowledge of disease prevalence in the data. The results from the four ophthalmologists were then fused by a senior ophthalmologist (who has more than ten years of experience in glaucoma), who checked the four markers and decided which of these markers should be retained to be averaged to produce the final reference coordinate.

\subsubsection{Optic Disc \& Cup Segmentation}
Similar to the previous task, the four ophthalmologists manually annotated the initial segmentation region of the optic cup and disc for each fundus image. The senior ophthalmologist then  examined the initial segmentations and selected the intersection of the annotated results of several ophthalmologists as the final reference mask.

\subsection{Challenge Evaluation}
\label{challenge_evaluation}
\subsubsection{Glaucoma Grading}
For each instance, the participants will predict normal, early-glaucoma or progressive-glaucoma. We use Cohen's kappa as an evaluation metric for this ordinal ternary classification problem. 
Since our categories are ordered, kappa is quadratically weighted to manifest the different extents of the error.
%\HBc{please rephrase. Don't know what sense the extent of error means}. 
The final score of glaucoma grading is represented as:
\begin{equation}
    S_{g} = 10 \times \kappa = 10 \ \times \frac{p^{o} - p^{e}}{1 - p^{e}},
\end{equation}
where $p^{o}$ is the accuracy, and $p^{e}$ is the probability of predicting the correct categories by chance.
%used to calculate the probabilities of randomly predicting the correct categories.

\subsubsection{Fovea Localization}
Fovea location is given by its X and Y coordinates. If the image does not contain a fovea, the estimated coordinate is set to be $(0,0)$. We use the average Euclidean distance between the estimated coordinates and the real coordinates as the evaluation criterion for this task. It is worth noting that the estimated and the ground-truth coordinate values are normalized according to the image size. The final score is based on the reciprocal of the average Euclidean distance (AED) value, and the denominator addition item is set to 0.1 to keep the score within 10 points:
\begin{equation}
	S_{f}=\frac{1}{AED+0.1}
\end{equation}

\subsubsection{Optic Disc \& Cup Segmentation}
The Dice coefficient was calculated as the segmentation evaluation metric in the GAMMA challenge:
%following the evaluation method of the previous challenges (\cite{orlandoREFUGEChallengeUnified2020a}) : 
\begin{equation}
Dice=\frac{2|A\cap B|}{|A|+|B|},
\end{equation}
where $|A|$ and $|B|$ represent the number of pixels of the prediction and ground truth, $|A \cap B|$ represents the number of pixels in the overlap between the prediction and ground truth. 
In addition, we used Mean Absolute Error (MAE) to measure the differences of the vertical cup-to-disc ratio (vCDR) between the predicted results and the ground truth. vCDR has a direct clinical relevance as it is a common measure used in ophthalmology and optometry to assess glaucoma progression. The vCDR is calculated as the ratio of the maximum vertical diameters of the optic cup and optic disc. Each team was ranked based on the three metrics of optic cup Dice coefficient, optic disc Dice coefficient, and MAE. The final weighted score for the optic disc \& cup segmentation task was as follows%\HBc{you should explain why the weights were so chosen}:
\begin{equation}
\begin{split}
S_{m} = & 0.25 \times Dice_{cup} \times 10 + 0.35 \times Dice_{cup} \times 10 \\
& + 0.4 \times \frac{1}{MAE+0.1}
\end{split}
\end{equation}
where, the weights were chosen consistent with the REFUGE Challenge (\cite{orlando2020refuge}). Because vCDR was calculated based on OC and OD segmentation results, the weight for vCDR metric had the highest value, and because the OD region could limit the OC region, the metric weight for OD segmentation was set higher than that for OC segmentation.
% In the end, the total scores of each team in the preliminary or final was calculated with the three competition tasks:
% \begin{equation}
% \begin{split}
% Score = & 0.4 \times Score_{task1} + 0.3 \times Score_{task2} \\
% & + 0.3 \times Score_{task3}
% \end{split}
%\end{equation}

\begin{table}[h]
\caption{Performance of the baselines for glaucoma grading. The results are shown in a format of mean(\%) $\pm$ standard deviation(\%). We run each method five times to calculate mean and standard deviation.}
\centering
\resizebox{\textwidth}{!}{%
\begin{tabular}{c|ccccc}
\hline
&\begin{tabular}[c]{@{}c@{}}Color\\ fundus\\ photography\end{tabular} & \begin{tabular}[c]{@{}c@{}}3D\\ OCT\end{tabular} & \begin{tabular}[c]{@{}c@{}}Disc\\ region\end{tabular} & \begin{tabular}[c]{@{}c@{}}Ordinal\\ regression\end{tabular} & Kappa \\ \hline
       \multirow{4}{1in}{Single-modality} & \checkmark         &    &        &  & 67.3$\pm$2.1 \\
       &  &   \checkmark &   &  & 57.5$\pm$3.6 \\
       &  \checkmark        &                                                  &                                        \checkmark               &                                                               & 67.7$\pm$1.8  \\ 
        & &  \checkmark  &   \checkmark    &    & 73.2$\pm$1.2 \\ \hline
       \multirow{4}{1in}{Multi-modality} &                                                   \checkmark                 &                         \checkmark                         &                                                       &                                                               & 70.2$\pm$0.9  \\

        & \checkmark    &                       \checkmark                           &                   \checkmark                                    &                                                               & 77.0$\pm$0.7 \\
        & \checkmark     &                   \checkmark                               &                                                       &                                      \checkmark                         & 76.8$\pm$0.6 \\
        & \checkmark      &                        \checkmark                          &                     \checkmark                                  &                                  \checkmark                             & 81.2$\pm$0.3 \\ \hline
\end{tabular}%
}
\label{table:before}
\end{table}

\begin{figure}[t]
\centering
\includegraphics[width=1\linewidth]{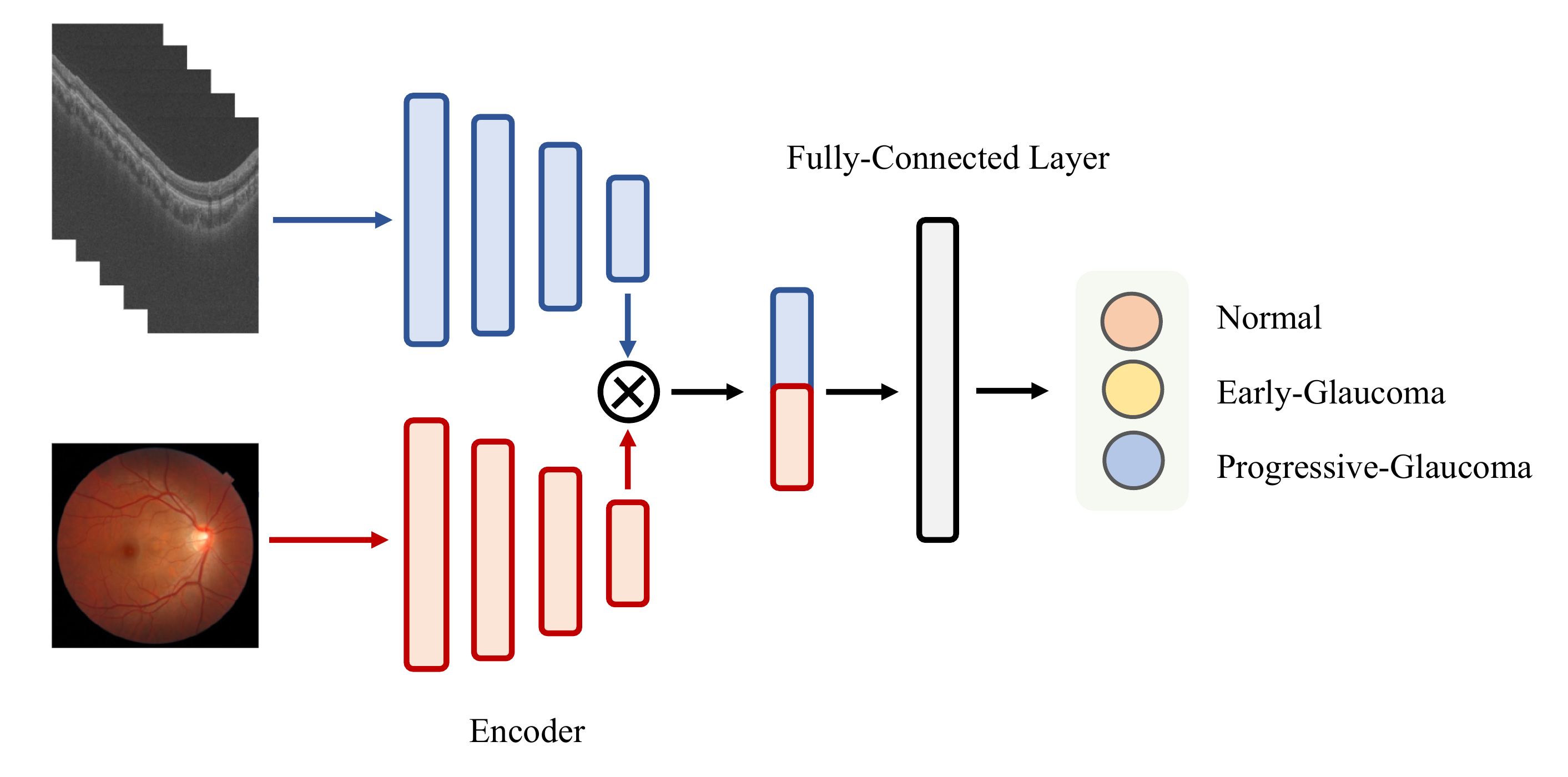}
\caption{Dual-branch network architecture for glaucoma grading. Blue blocks denote the OCT network branch. Red blocks denote the fundus network branch. The features of two branches are concatenated for the final classification.}
\label{fig:dualbranch}
\end{figure}

\subsection{Baseline method}\label{sec:baseline}
Before the challenge start, we provided a method to serve as a baseline implementation and performance for the challenge. As deep end-to-end learning has been proved to be widely effective for the biology and medical image analysis (\cite{zhang2020high, ge2022srs})

A simple dual-branch network was used to learn glaucoma grading from fundus images and 3D OCT volumes in an end-to-end manner. An illustration of the architecture is shown in Figure \ref{fig:dualbranch}. Specifically, two CNN-based encoders are used to extract the features from fundus images and OCT volume, respectively. Two encoders are implemented following ResNet34 (\cite{he2016deep}) with the same architecture except for the first convolutional layer. In the fundus branch, the input channel of the first convolutional layer is set as 3, and in the OCT branch, it is set as 256. The encoded features of the fundus branch and OCT branch are concatenated and classified by a fully connected layer. The model is supervised by cross-entropy loss function in the training stage. We trained it on the training dataset, evaluated it on the preliminary stage data and reported its performance on the final test data. We input the fundus images with resolution $256 \times 256$, and the OCT images with resolution $512 \times 512$. We train the networks using Adam optimizer (\cite{kingma2014adam}) with batch size 4. More details of the baseline can be found in (\cite{fang2021multi}). The code of the baseline is released at \url{https://aistudio.baidu.com/aistudio/projectdetail/1948228}.
%$$\HBc{So not on preliminary stage data?}.

In the clinic, ophthalmologists use a combination of fundus photographs and OCT volumes for a more accurate and reliable diagnosis. We find that this approach is still applicable in deep learning-based computer-aided glaucoma diagnosis. We compare the performance of the single fundus branch, single OCT branch, and dual-branch baseline in Table \ref{table:before}. Of note, only the basic dual-branch model was released as the baseline to the participants (the fifth row in Table~\ref{table:before}). From the table, one can observe that the dual-branch model outperforms the single branch one by a large margin with less variances. This indicates that despite the simple multimodal fusion strategy we adopted, multi-modal images can improve the glaucoma grading performance better than either of the modalities alone. This motivated us to hold the GAMMA Challenge to encourage the further exploration of advanced machine learning methods on this multimodal fusion task.

During the implementation of the baseline, we identified some techniques that were found particularly useful to obtain good performance on the task (\cite{fang2021multi}). The first is to utilize the local information of optic disc. Clinically, glaucoma leads to lesions in the optic disc region, such as cup-disc ratio enlargement and optic disc hemorrhage (\cite{orlando2020refuge}). Thus, we cropped the optic disc region of fundus images as the network's input to make the network focus on the optic disc and cup. The optic disc region is obtained through pre-trained optic disc segmentation network. According to the results in Table \ref{table:before}, the local information extraction gains 7.2\% improvement on mean kappa and 0.2\% on standard deviation compared with the baseline.
%\HBc{you also had disc region in OCT according to the Table (row 4)? how was ONH identified there}.
 
We also note that glaucoma grading is actually an ordinal classification task. The three classifications: normal, early-glaucoma, and progressive-glaucoma, are the deterioration of glaucoma. Thus, in the training process, the loss should be smaller if the prediction is closer to the ground-truth. For example, predicting the early-glaucoma as normal should be considered as a smaller error than predicting the progressive-glaucoma as normal. Therefore, we adopted ordinal regression strategy (\cite{niu2016ordinal}) to perform two binary classifications, respectively. In this case, a severe error will be double-penalized by both of the classifiers. Specifically, the first classier divides the sample into 0 and 1, that is, to classify whether the input image is a glaucoma sample. The second classier divides the sample into 0 and 1 to identify the input image as progressive-glaucoma or early-glaucoma. The labels of the original triple classification task were converted according to the two binary classification tasks, that is, the labels of the normal samples were changed to (0,0), the labels of the early-glaucoma samples were changed to (1,0), and those of the progressive-glaucoma samples were changed to (1,1). The loss function used in the training processes was the sum of the two binary cross-entropy losses. According to the results in Table~\ref{table:before}, ordinal regression independently resulted in an average 4.5\% improvement of the models.

\section{Methods of participating teams}
The methodology applied by the ten top performing teams in the GAMMA Challenge is summarized in Table \ref{table:all}. In this section, we introduce their methods in the aspects of data preprocessing, architecture and ensembling strategy.
\begin{table}[!htbp] 
\centering
\caption{Summary of the ten top performing glaucoma grading methods in the GAMMA Challenge.}
\resizebox{0.9\textwidth}{!}{%
\begin{tabular}{c|c|c|c|c}
\hline
Team        & Architecture                                                                                                    & Preprocessing                                                                                                                                                                                                                                                                                                                                                                                                                                                                & Ensemble                                                                                                                                                                                                                                                                                                                       & Method                                                                                                                                                                                                                                                                                                                     \\ \hline
SmartDSP (\cite{cai2022corolla})    & \begin{tabular}[c]{@{}c@{}}Dual-branch ResNet (\cite{he2016deep}) \end{tabular}                                                                                                & \begin{tabular}[c]{@{}c@{}}Fundus: Add Gaussian noise\\ Resize to 512\texttimes 512\\ OCT: Crop height to 150-662\\ Resize to 512\texttimes512\\  Default Data Augmentation\end{tabular}                                                                                                                                                                                                                                                                                           & \begin{tabular}[c]{@{}c@{}}Pick 3 models with \\ best accuracy on normal,\\  early and progressive cases, \\ respectively. Predict the \\ results by different thresholds.\\  Ensemble the results by\\  the priorities of early, \\  progressive and normal.\end{tabular} & \begin{tabular}[c]{@{}c@{}}Extract the features of \\ fundus images and OCT \\ volumes by two encoders.\\ Concatenate the encoded\\ features for the classification.\end{tabular}                                                                                                                                                   \\ \hline
VoxelCloud  & \begin{tabular}[c]{@{}c@{}}Dual-branch Network \\ implemented by\\ 3D EfficientNet and EfficientNet\\ (\cite{tan2019efficientnet})\end{tabular} & \begin{tabular}[c]{@{}c@{}}Fundus: Crop Black Margin\\ Resize to 512\texttimes512\\  OCT: Resize to 256\texttimes256\\ Downsample channels to 128\\  Default Data Augmentation\end{tabular}                                                                                                                                                                                                                                                                                                                               & \begin{tabular}[c]{@{}c@{}}Pick 5 best models on 5 \\ different validation folds.\\ Ensemble the results \\ by taking the average.\end{tabular}                                                                                                                                                                                 & \begin{tabular}[c]{@{}c@{}}Extract the features of \\ fundus images by EfficientNet.\\ Extract the features of \\ OCT volumes by 3D-EfficientNet.\\ Concatenate the encoded\\ features for the classification.\end{tabular}                                                                                                         \\ \hline
EyeStar     & \begin{tabular}[c]{@{}c@{}}Dual-branch Network\\ implemented by \\Swin Transformer (\cite{liu2021swin})\\ and DENet (\cite{Fu2018DiscAware}) \end{tabular}              & \begin{tabular}[c]{@{}c@{}}Fundus: Crop to optic disc region\\ by pretrained segmentation network\\  OCT: Randomly pick ten consecutive \\slices betwern 113-153 channels\\  Default  Data Augmentation\end{tabular}                                                                                                                                                                                                                                                   & \begin{tabular}[c]{@{}c@{}}During the testing process,  \\ successively feed 30 \\ groups of 10\\ consecutive OCT slices\\  into the network. \\ Taking the average of the 30 \\ predictions as the final predictions\end{tabular}                                                                                                   & \begin{tabular}[c]{@{}c@{}}Extract the features of \\ fundus images by fundus disc-aware \\ ensemble network. \\ Extract the features of \\ OCT volumes by ResNet.\\ Concatenate the encoded\\ features for the classification.\end{tabular}                                                                                  \\ \hline
HZL         & \begin{tabular}[c]{@{}c@{}}UNet (\cite{ronneberger2015u}) with\\ EfficientNet Backbone\end{tabular}                                      & \begin{tabular}[c]{@{}c@{}}Fundus: Resize to 1024\texttimes1024\\  OCT: Resize to 1024\texttimes1024\\  Default  Data Augmentation\end{tabular}                                                                                                                                                                                                                                                                                                                                                            & \begin{tabular}[c]{@{}c@{}}Pick 5 best models on 5 \\ different validation folds.\\ Ensemble the results \\ by taking the average.\end{tabular}                                                                                                                                                                                 & \begin{tabular}[c]{@{}c@{}}Design a multi-task UNet\\ to jointly learn glaucoma grading, \\ optic disc \& cup segmentation \\ and fovea localization. \\ The embedding of the UNet encoder\\ is discriminated by a full connected\\ layer for glaucoma grading.\end{tabular}                                                       \\ \hline
MedIPBIT    & Dual-branch EfficientNet                                                                                         & \begin{tabular}[c]{@{}c@{}}Fundus: Crop to optic disc region \\ by pretrained segmentation network.\\ Resize to 128\texttimes128\\  OCT: Crop the Black Background \\ by gradient detector \\Resize to 128\texttimes128\\  Default Data Augmentation\end{tabular}                                                                                                                                                                                                                                                                                   & \begin{tabular}[c]{@{}c@{}}Split the dataset for \\ training and validation  \\ by three different strategies.\\ Pick 2 best performing models \\ in each split \\to get a total of 6 models.\\ Ensemble the results of 6 models \\ by averaging.\end{tabular}                                                                                & \begin{tabular}[c]{@{}c@{}}Extract the features of \\ fundus images and OCT  \\ volumes by two encoders.\\ Concatenate the encoded\\ features for the classification.\end{tabular}                                                                                                                                                   \\ \hline
IBME        & Dual-branch ResNet                                                                                               & \begin{tabular}[c]{@{}c@{}}Fundus: Resize to 256\texttimes256\\  OCT: Resize to 512\texttimes512\\  Default  Data Augmentation\end{tabular}                                                                                                                                                                                                                                                                                                                                                              &                                                                                                                                                                                                                                                                                                                               & \begin{tabular}[c]{@{}c@{}}Extract the features of \\ fundus images and OCT \\ volumes by two encoders.\\ Concatenate the encoded\\ features for the classification.\end{tabular}                                                                                                                                                   \\ \hline
WZMedTech   & Dual-branch ResNet                                                                                               & \begin{tabular}[c]{@{}c@{}}Fundus: Resize to 512\texttimes512 \\ Default  Data Augmentation + Image Jitter\\  OCT: Resize to 256\texttimes256\end{tabular}                                                                                                                                                                                                                                                                                                                                                 & \begin{tabular}[c]{@{}c@{}}Pick the first and the \\second best model.\\ Predict as normal when \\ both models predicted \\ the case as normal. \\ Use the output of the OCT branch \\ of the second best model when \\ either of the two \\ models predicts glaucoma. \end{tabular}                                              & \begin{tabular}[c]{@{}c@{}}Predict glaucoma grading \\ based on fundus images\\ and OCT volume by two networks.\\ Take the average of \\ the two networks' results.\end{tabular}                                                                                                                                               \\ \hline
DIAGNOS-ETS & \begin{tabular}[c]{@{}c@{}}Dual-branch Network \\ implemented by\\ 3D ResNet (\cite{tran2015learning}) and ResNet\end{tabular}             & \begin{tabular}[c]{@{}c@{}}Fundus: Resize with the shorter spatial \\ side randomly sampled in 224 to 480 \\ and randomly crop to 224\texttimes224\\  OCT: Downsample channels to 16 \\ Randomly pick one slice in training \\ Pick specific slices in the inference\\ Crop width to 224-480  \\ Resize the original images with \\ shorter spatial side randomly \\ sampled in range 256-480 \\  Default Data Augmentation\end{tabular} & \begin{tabular}[c]{@{}c@{}}Ensemble multi-scale \\ prediction by averaging them \\ with temperature scaling\end{tabular}                                                                                                                                                                                                    & \begin{tabular}[c]{@{}c@{}}Extract the features of \\ fundus image and OCT volume\\ by ResNet and 3D ResNet, \\ respectively.\\ Concatenate the encoded\\ features for the classification.\\ During training, the encoded features of \\ two networks are aligned by \\ minimizing the KL divergence\end{tabular}                             \\ \hline
MedICAL     & Dual-branch EfficientNet                                                                                         & \begin{tabular}[c]{@{}c@{}}Fundus: Resize 1024\texttimes1024 \\ Enhanced by optic disc and cup mask\\  OCT: Transfer to Retina Tickness Heatmap \\ Resize 400\texttimes400\\  Default  Data Augmentation\end{tabular}                                                                                                                                                                                                                                                                            & \begin{tabular}[c]{@{}c@{}}Take the average \\ of multiple trained models\end{tabular}                                                                                                                                                                                                                                              & \begin{tabular}[c]{@{}c@{}}Extract the features of \\ fundus images and OCT \\ volumes by two encoders.\\ Concatenate the encoded\\ features for the classification.\end{tabular}                                                                                                                                                   \\ \hline
FATRI-AI    & EfficientNet                                                                                                    & \begin{tabular}[c]{@{}c@{}}Fundus: Crop Black Margin\\ Resize 224\texttimes224\\  OCT: Random pick 3 slices\\ Resize 224\texttimes224\\  Default Data Augmentation\end{tabular}                                                                                                                                                                                                                                                                                                                      &\begin{tabular}[c]{@{}c@{}}Stack two models,\\ output with confidence \\ $>0.7$ in the first model\\ is used as pseudo labels \\ to train the second model.\end{tabular}                                                                                                                                                                                                                                                                                                                          & \begin{tabular}[c]{@{}c@{}}Concatenate fundus image \\  and OCT volume as the input\\ to a single network.\\ The network predicts\\ the probability of each class.\end{tabular} \\ \hline
\end{tabular}%
}
\label{table:all}
\end{table}

\subsection{Data Preprocessing}
% Data preprocessing typically includes cropping, resizing and data augmentation. 
%In the experiment, we do find that to cropping image to the region of interest and keeping a high resolution will take marginal improvement. 
In the baseline implementation, we provided a default data augmentation implemented by some commonly used data augmentation techniques, including random crop, random flip and random rotation. Most of the teams used this default augmentation for data preprocessing. 

\begin{figure}[h]
\centering
\includegraphics[width=1\linewidth]{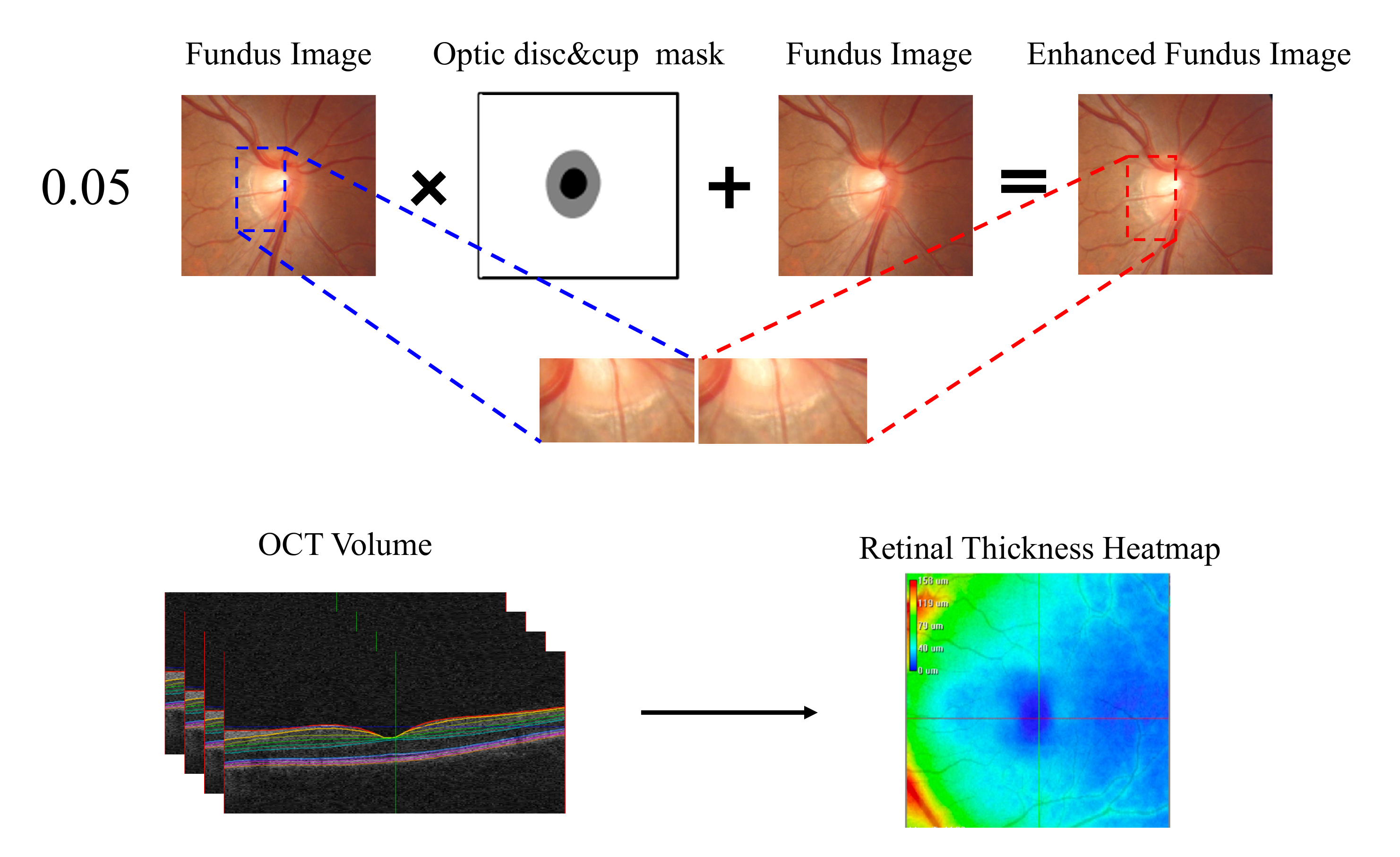}
\caption{Data preprocessing of MedICAL. They enhanced fundus images by OD/OC mask and transfer the 3D OCT volume as 2D retinal thickness map.}
\label{fig:pre}
\end{figure}

Besides the standard data augmentation, during training, DIAGNOS-ETS augments the input samples by rescaling with the shorter spatial side randomly sampled in a range of 224 to 480, and cropping with size of $224 \times 224$. In the test phase, they do test-time augmentation for multi-scale ensemble. Inputs are
spatially resized such that the shorter sides are 224, 256, 384, 480 respectively for each model, and all cropped to $224 \times 224$.Then they adopted ensemble over multi-scale results for the prediction. MedIPBIT cropped the fundus images to the optic disc region. In the training stage, they used the optic disc mask provided in GAMMA dataset for this cropping. In the inference stage, they used instead the masks estimated by the pre-trained segmentation network. The segmentation networks were trained on the auxiliary tasks on GAMMA dataset. Besides MedIPBIT, MedICAL also utilized OD/OC mask for data preprocessing. They enhanced their fundus image by OD/OC mask. Specifically, OD/OC region of the original image will be multiplied by a factor of 0.05 and added to the original image. MedICAL also transferred the 3D OCT volume to 2D retinal thickness heatmap by Iowa Reference Algorithm (\cite{rosenthal2016ophthalvis}). An illustration of their process is shown in Figure \ref{fig:pre}. 

% In the experiment, the performance changes caused by different data preprocessing methods are not obvious

\begin{figure}[!t]
\centering
\includegraphics[width=1\linewidth]{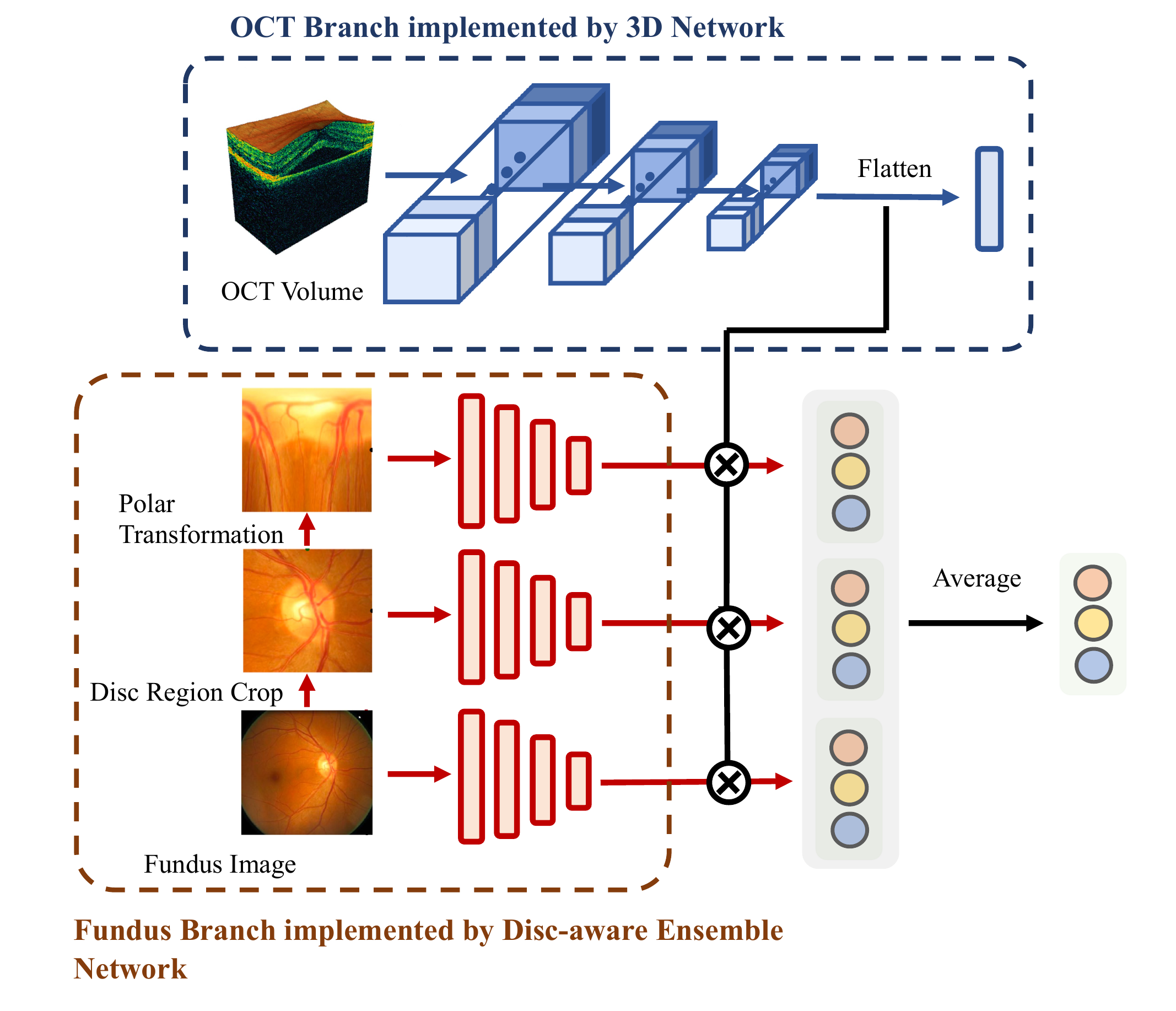}
\caption{Illustration of OCT 3D Network branch and fundus DENet branch. 3D Network is adopted by team VoxelCloud and team DIAGNOS-ETS. DENet is adopted by team EyeStar. For the OCT 3D Network branch, the encoded feature is flattened and concatenated with that of the fundus branch. For the fundus Disc-aware ensemble branch, the features of three subbranches are concatenated with OCT features for the classification, respectively. The final prediction is the average of the three subbranches. }
\label{fig:3de}
\end{figure}

\subsection{Architecture}
For the fundus \& OCT-based glaucoma grading, almost all the teams adopted dual-branch network structure. Analogously to the baseline method, two branches extract the features of fundus images and OCT volumes. The encoded features are then concatenated for the classification. Unlike this strategy, FATRI-AI used a single network inputted by concatenated fundus image and OCT volume. Besides FATRI-AI, HZL also used a single branch of network. They proposed a multi-task UNet network to jointly learn the glaucoma grading, optic disc \& cup segmentation and fovea localization. The glaucoma grading head is attached to the UNet encoder, while the segmentation head and localization head are attached to the UNet decoder. Through the multi-task learning strategy, the correlated features of different tasks will be enhanced and thus improve the performance of all the tasks. 

Although most of the teams adopted a dual-branch network architecture, their implementations varied greatly. VoxelCloud and DIAGNOS-ETS adopted 3D Network  (\cite{tran2015learning}) in OCT branch to extract the features from 3D OCT volume. EyeStar adopted fundus Disc-aware Ensemble Network (DENet) (\cite{Fu2018DiscAware}) in fundus branch. Fundus disc-aware ensemble network uses three networks to respectively process the raw fundus image, optic disc region of the fundus image, and polar transformed optic disc region. The predictions of three networks are combined to obtain the final prediction. An illustration of 3D network and DENet is shown in Figure \ref{fig:3de}. WZMedTech used two independent networks to predict glaucoma grades based on fundus image and OCT volume, respectively. The final result is the average of the two predictions.

Regarding the supervision signal, most teams applied cross-entropy loss. DIAGNOS-ETS has an extra loss to align the fundus feature and OCT feature. Toward that end, they minimize the Kullback–Leibler (KL) divergence between these two encoded features. Instead of supervising the integrated features of two modalities, EyeStar and WZMedTech supervised the two branches independently. They took the average of the independent predictions as the final result. In the ablation experiments, we did not observe differences between these supervision strategies.

\subsection{Ensemble strategy}
Ensembling can substantially improve the quantitative result of glaucoma grading. A basic idea is to pick the best models on different validation folds and take the average of the results. Teams VoxelCloud, HZL, MedIPBIT, MedICAL adopted this strategy. 

A unique ensemble strategy adopted in GAMMA Challenge is to exploit the ordinal nature of class labels for ensembling. %Before the challenge, we have found this ordered nature of the class lables, i.e., identifying progressive-glaucoma as normal is a more serious error than identifying early glaucoma as normal (\cite{fang2021multi}). %Therefore, even if the number of errors is the same, the prediction closer to the gold standard will gain higher performance measured by weighted kappa evaluation metric.
Separating the triple-classification problem into two binary-classification ones can help to improve the results. Both SmartDSP and WZMedTech adopted a similar idea for their ensembling strategy. WZMedTech discriminated early/progressive cases based on the dual-model agreed glaucoma cases. They double-checked the normal cases by two different models, i.e., first discriminated the normal/glaucoma cases, then classified progressive/early by the second model on predicted glaucoma cases. SmartDSP followed the same high-level idea, but adopted a more sophisticated strategy. They first picked three models with the best accuracy on normal, early, and progressive cases, respectively. Then they discriminated the progressive cases by thresholding the progressive model with 0.6, discriminated the cases as glaucoma by thresholding the normal model with 0.5, discriminated early-glaucoma by thresholding the early model with 0.9. The samples rejected by all three models will be classified as early-glaucoma by default. The pseudo code of this process is shown in Algorithm \ref{alg:cap}.
%In this way, the predictions will be ordered and thus perform better on weighted kappa metric.
\begin{algorithm}[t]
\caption{Ensembling strategy of SmartDSP}\label{alg:cap}
\begin{algorithmic}[1]
% \setalglineno{1}
\State Train $k$ models, and pick three models with the best accuracy on normal, early, and progressive cases, which are denoted as $M_{n}$, $M_{e}$, and $M_{p}$, respectively.
% \setalglineno{2}
\For {each sample $x$ in dataset $X$}
% \setalglineno{3}
    \State $x \gets early-glaucoma$
% \setalglineno{4}
\If{$M_{p}(x) > 0.6$}
% \setalglineno{5}
   \State $x \gets progressive-glaucoma$ 
% \setalglineno{6}
\EndIf
% \setalglineno{7}
\If{$M_{n}(x) < 0.5$}
% \setalglineno{8}
   \State $x \gets normal$
% \setalglineno{9}
\EndIf
% \setalglineno{10}
\If{$M_{e}(x) > 0.9$}
% \setalglineno{11}
   \State $x \gets early-glaucoma$
% \setalglineno{12}
\EndIf
% \setalglineno{13}
\EndFor
\end{algorithmic}
\end{algorithm}
Besides these ensembling strategies, DIAGNOS-ETS rescaled the input images to different sizes in the inference stage, and combined the multi-scale predictions by averaging them with temperature scaling. Specifically, they combines the multi-scale results through:
\begin{equation}
    p = \sum_{i=1}^{N} \frac{p_{i}^{t}}{N},
\end{equation}
where $p_{i}$ are the multi-scale predictions, $p$ is the final prediction, $N$ is the number of scales, $t$ is a learned scalar parameter. FATRI-AI stacked two models, where the instances with high confidence in the first model (over 0.7) were used as pseudo labels to train the second model.

\section{Results}\label{results}
\subsection{Challenge Results}
The top ten teams ranked by glaucoma grading score are SmartDSP, Voxelcloud, EyeStar, HZL, MedIPBIT, IBME, WZMedTech, DIAGNOS-ETS, MedICAL, and FATRI\_AI. The quantitative scores of the glaucoma grading task measured by kappa are shown in Table \ref{tab:rank}. We reported their performances in the preliminary stage (evaluation on  preliminary set) and the final stage (evaluation on final test set). Comparing the ranking in the preliminary stage with that of the final stage, we can see SmartDSP, Voxelcloud, EyeStar, HZL, IBME all keep or raise the rankings on the test dataset, indicating they are more robust than the other methods. 
%We note that the methods which are more robust also rank high in the GAMMA Challenge, i.e., the top-4 teams: SmartDSP, Voxelcloud, EyeStar, HZL show more robust performance. 
The teams ranked lower are generally caused by the worse generalization capability. In particular, for MedIPBIT, IBME, WZMedTech, DIAGNOS-ETS, and MedICAL, we can see a dramatic decrease of the performance on the final test set.

\begin{figure*}[h]
\centering
\includegraphics[width=\textwidth]{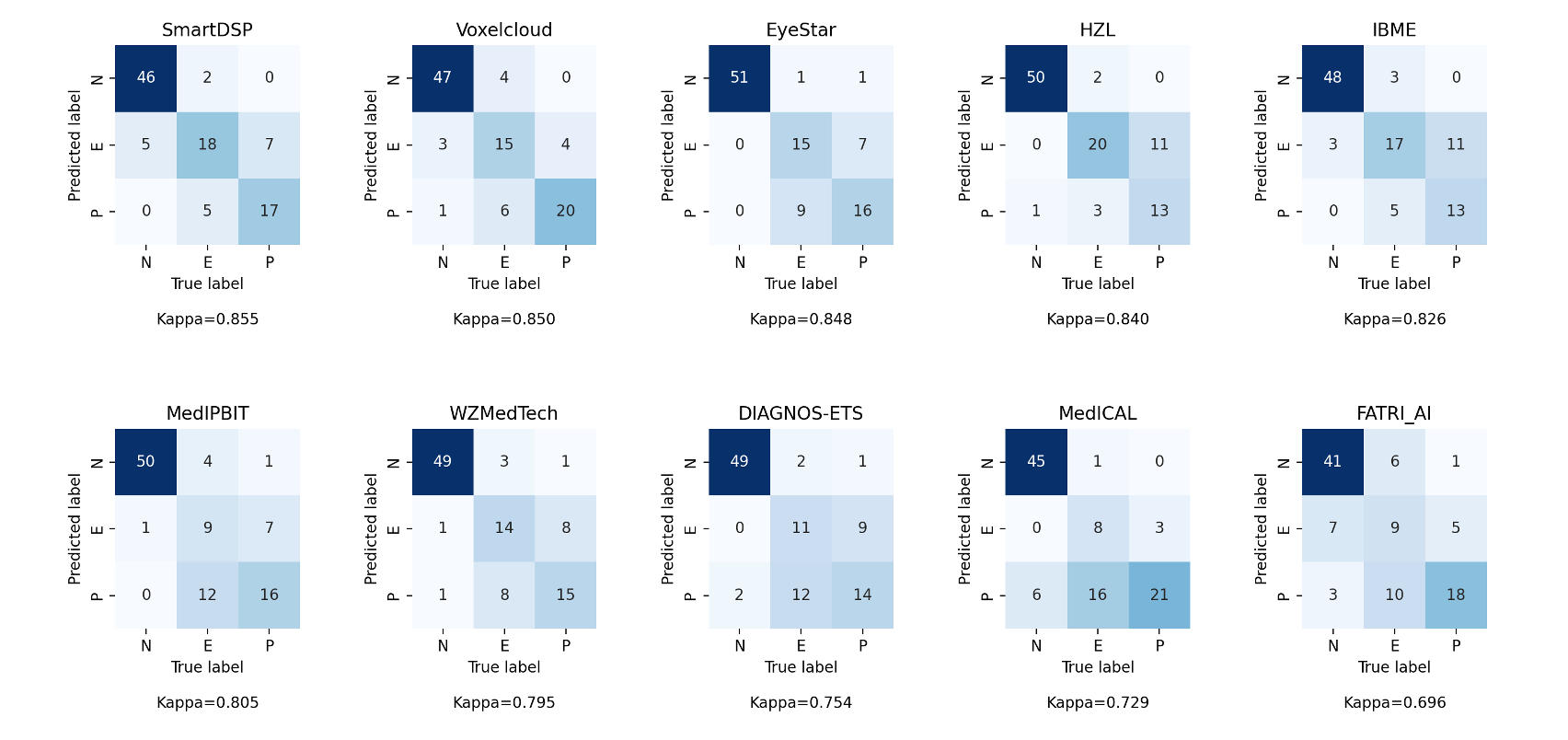}
\caption{Glaucoma grading confusion matrix of each team. N, E, P denote normal, early-glaucoma and progressive-glaucoma respectively.}
\label{fig:gamma_cf}
\end{figure*}

The confusion matrices calculated on the test set are shown in Figure~\ref{fig:gamma_cf}. We note that methods achieved similar performance in the prediction of normal/glaucoma. The error of predicting glaucoma as normal is generally in 4\% to 8\% range. This rate is lower than the reported misdiagnosed rate of junior ophthalmologists (\cite{trobe1980optic}), indicating the clinical application potential of the models.

Different approaches widened the gap in the performance of early/progressive-glaucoma classification. Teams ranked higher generally achieved better performance on both the early-glaucoma accuracy and progressive-glaucoma accuracy. It is also worth noting that the accuracy of early glaucoma and progressive glaucoma has different significance in clinical scenarios. In clinical scenarios, predicting progressive-glaucoma as early-glaucoma is generally more undesirable than the other way around. Thus, among models with similar overall performance, those with higher progressive-glaucoma accuracy will be a better choice in clinical practice.

\begin{table}[h]
\caption{Glaucoma grading results in the GAMMA Challenge. Kappa(\%) is calculated to measure the performances. Teams are ranked by the overall score. \textcolor{red}{Red} and \textcolor{blue}{blue} denote the rise and fall of the rankings, respectively, while \textcolor{gray}{Gray} denotes no change in the ranking between the preliminary and the final test stage.}
\centering
\begin{tabular}{c|c|cc}
\hline
\multirow{2}{*}{Rank} & \multirow{2}{*}{Team} & \multirow{2}{*}{Preliminary} & \multirow{2}{*}{Final}  \\
                      &                       &                             &                                                \\ \hline
1                     & SmartDSP              & 93.38 \textcolor{gray}{(1)}                       & 85.49                % & 87.86                    
\\
2                     & VoxelCloud            & 90.71 \textcolor{red}{(6)}                       & 85.00                 % & 86.72                   
\\
3                     & EyeStar               & 88.28 \textcolor{red}{(7)}                       & 84.77                 % & 85.82                  
\\
4                     & HZL                   & 89.89 \textcolor{red}{(8)}                       & 84.01                 % & 85.78                  
\\
5                     & IBME                  & 87.60 \textcolor{red}{(9)}                       & 82.56                %  & 84.07                   
\\
6                     & MedIPBIT              & 93.43 \textcolor{blue}{(2)}                       & 80.48                 % & 84.36                  
\\
7                     & WZMedTech             & 90.44 \textcolor{blue}{(5)}                       & 79.46                 % & 82.76                  
\\
8                     & DIAGNOS-ETS           & 91.70 \textcolor{blue}{(3)}                       & 75.36                %  & 80.26                  
\\
9                     & MedICAL               & 90.65 \textcolor{blue}{(4)}                       & 72.90                % & 78.23                  
\\
10                    & FATRI-AI              & 87.34 \textcolor{gray}{(10)}                       & 69.62                 % & 74.94                   
\\ \hline
\end{tabular}
\label{tab:rank}
\end{table}

To encourage the teams participate in all three tasks of the GAMMA challenge, the official final ranking is calculated with the scores from all three competition tasks:
\begin{equation}
\begin{split}
Score = & 0.4 \times Score_{g} + 0.3 \times Score_{f} \\
& + 0.3 \times Score_{m}.
\end{split}
\end{equation}
The published final ranking is shown in Table \ref{tab:final_rank}. The ranking of the auxiliary tasks is shown in the Appendix. The detailed leaderboards can also be accessed on the GAMMA challenge website at \url{https://aistudio.baidu.com/aistudio/competition/detail/90/0/leaderboard}.

\begin{table*}[h]
\centering
\caption{Final ranking of the GAMMA challenge.}
\resizebox{\textwidth}{!}{%
\begin{tabular}{c|c|c|c|c}
\hline
\multirow{2}{*}{Rank} & \multirow{2}{*}{Team} & \multirow{2}{*}{Member}                                                                                                     & \multirow{2}{*}{Institute}                                                                                                       & \multirow{2}{*}{Score} \\
                      &                       &                                                                                                                              &                                                                                                                                  &                        \\ \hline
1                     & SmartDSP              & \begin{tabular}[c]{@{}c@{}}Jiongcheng Li, Lexing Huang, Senlin Cai, \\ Yue Huang, Xinghao Ding\end{tabular}                  & Xiamen University                                                                                                                & 8.88892                \\
2                     & Voxelcloud            & \begin{tabular}[c]{@{}c@{}}Qinji Yu, Sifan Song, Kang Dang, Wenxiu \\ Shi, Jingqi Niu\end{tabular}                           & \begin{tabular}[c]{@{}c@{}}Shanghai Jiao Tong University; Xi'an \\ Jiaotong-Liverpool University; VoxelCloud \\ Inc.\end{tabular} & 8.83127                \\
3                     & EyeStar               & \begin{tabular}[c]{@{}c@{}}Xinxing Xu, Shaohua Li, Xiaofeng Lei, Yanyu \\ Xu, Yong Liu\end{tabular}                          & Institute of High Performance Computing, ASTAR                                                                                   & 8.72345                \\
4                     & IBME                  & Wensai Wang, Lingxiao Wang                                                                                                   & \begin{tabular}[c]{@{}c@{}}Chinese Academy of Medical Sciences and \\ Peking Union Medical College\end{tabular}                  & 8.70783                \\
5                     & MedIPBIT              & \begin{tabular}[c]{@{}c@{}}Shuai Lu, Zeheng Li,Hang Tian,Shengzhu \\ Yang,Jiapeng Wu\end{tabular}                            & Beijing Institute of Technology                                                                                                  & 8.70561                \\
6                     & HZL                   & Shihua Huang, Zhichao Lu                                                                                                     & \begin{tabular}[c]{@{}c@{}}Hong Kong Polytechnic University; \\ Southern University of Science and \\ Technology\end{tabular}    & 8.68781                \\
7                     & WZMedTech             & \begin{tabular}[c]{@{}c@{}}Chubin Ou, Xifei Wei, Yong Peng, \\ Zhongrong Ye\end{tabular}                                     & \begin{tabular}[c]{@{}c@{}}Southern Medical University; Tianjin \\ Medical University; Xinjiang University\end{tabular}          & 8.65384                \\
8                     & DIAGNOS-ETS           & \begin{tabular}[c]{@{}c@{}}Adrian Galdran,Bingyuan Liu,José \\ Dolz,Waziha Kabir,Riadh Kobbi,Ismail Ben \\ Ayed\end{tabular} & ETS Montreal; DIAGNOS Inc.                                                                                                       & 8.59884                \\
9                     & MedICAL               & Li Lin, Huaqing He, Zhiyuan Cai                                                                                              & \begin{tabular}[c]{@{}c@{}}Southern University of Science and \\ Technology\end{tabular}                                         & 8.43841                \\
10                    & FATRI\_AI             & \begin{tabular}[c]{@{}c@{}}Qiang Zhou, Hu Qiang, Cheng Zheng, \\ Tieshan Liu, Dongsheng Lu, Xinting Xiao\end{tabular}        & Suixin (Shanghai) Technology Co., LTD.                                                                                           & 8.27601                \\ \hline
\end{tabular}%
}\label{tab:final_rank}
\end{table*}

\subsection{Methodological Findings}\label{sec:finding}
In this section, we draw the key methodological findings by doing the ablation study on the techniques proposed in the GAMMA challenge. A brief conclusion is that a 3D Net \& DENet dual branch architecture with ordinal ensemble strategy performs best on this task. The focus on the OD/OC region also helps to improve the glaucoma grading. The detailed analysis and discussion are as follow. 
%The comparison and analysis will help verify the effective techniques for the task and contribute to the future development of novel methods.

\subsubsection{Ablation study on architectures in GAMMA}
In order to fairly verify the effectiveness of proposed architectures, we did an ablation study utilizing our baseline implementation as reference. We kept everything the same as the baseline and only changed the architectures. The quantitative results are shown in Table \ref{tab:arch}. We measure the results by the overall kappa and also the accuracy value of each class. N-Acc, E-Acc and P-Acc denotes the accuracy values of normal, early-glaucoma and progressive-glaucoma, respectively. G-Acc denotes the glaucoma accuracy value of both early-glaucoma and progressive-glaucoma classes.

\begin{table}[h]
\centering
\caption{Comparison of the network architectures in the GAMMA Challenge. 'DualRes' denotes the dual-branch ResNet architecture adopted by SmartDSP, MedIPBIT, IBME, WZMedTech and MedICAL. 'Res-3D'
denotes a dual-branch ResNet architecture with a 3D-ResNet OCT branch and a standard fundus branch, which VoxelCloud and DIAGNOS-ETS adopt. 'Res-DEN' denotes a dual-branch ResNet architecture with DENet fundus branch and standard OCT branch, which EyeStar adopts. 'SinCat' and 'SinMulti' denote the single network inputted by fundus \& OCT concatenation and multi-task learning strategy adopted by FATRI-AI and HZL, respectively. The results are shown in a format of mean(\%) $\pm$ standard deviation(\%). We run each method five times to calculate mean and standard deviation.}
\resizebox{0.8\textwidth}{!}{%
\begin{tabular}{c|ccccc}
\hline
                              & N-Acc & E-Acc & P-Acc & G-Acc & Kappa \\ \hline
DualRes                       & 84.31$\pm1.77$ & 27.20$\pm2.26$ & 71.66$\pm2.33$ & 42.04$\pm1.79$ & 70.26$\pm0.94$ \\
Res-3D                        & 82.74$\pm1.27$ & 24.26$\pm2.02$ & 73.32$\pm1.21$ & 52.18$\pm1.07$ & 73.81$\pm0.48$ \\
Res-DEN                       & 94.63$\pm1.02$ & 24.21$\pm1.21$ & 76.75$\pm1.62$ & 48.13$\pm0.95$ & 76.82$\pm0.41$ \\
SinCat                        & 74.31$\pm1.23$ & 33.60$\pm3.60$ & 52.50$\pm2.85$ & 42.86$\pm2.22$ & 61.31$\pm1.82$ \\
\multicolumn{1}{l|}{SinMulti} & 89.02$\pm2.24$ & 20.80$\pm3.18$ & 67.21$\pm2.53$ & 51.46$\pm1.91$ & 75.31$\pm1.09$ \\ \hline
3D-DEN                        & 97.88$\pm0.91$ & 33.12$\pm1.76$ & 54.56$\pm1.86$ & 43.14$\pm0.92$ & 79.55$\pm0.21$ \\ \hline
\end{tabular}
\label{tab:arch}
}
\end{table}

From Table \ref{tab:arch}, we can see, first, the awareness of optic disc region is helpful for glaucoma grading. Res-DEN and SinMulti utilized the optic disc \& cup segmentation mask, and they achieved higher and steadier performance. In particular, Res-DEN achieves a much better overall performance than standard DualRes (increases a 6.42\% on mean kappa and decreases a 0.53\% on standard deviation), indicating DENet is a better choice than the standard network for the fundus branch. In addition, Res-3D outperforms DualRes by a 3.41\% on mean kappa and a 0.46\% on kappa standard deviation, indicating 3D neural network works better than the standard network as an OCT branch. We also tried to combine the two advantages, by adopting 3D network on OCT branch and adopting DENet on fundus branch. The results are denoted as 3D-DEN in Table \ref{tab:arch}. We can see that the combined architecture outperforms both approaches. Specifically, it outperforms Res-3D by a 5.74\% on mean kappa, outperforms Res-DEN by a 2.73\% on mean kappa, and outperforms basic DualRes by an outstanding 9.15\% on mean kappa with the lowest standard deviation among all methods. In conclusion, in terms of the architecture, a 3D neural network OCT branch with a DENet fundus branch is suggested for multi-modal glaucoma grading.

\subsubsection{Ablation study on ensemble strategies in GAMMA}
We also performed the ablation study on the ensemble strategies proposed by top ten teams. The quantitative results are shown in Table \ref{tab:ensemble}. 
%\HBc{I think this would be more fitting for the Table caption rather than here}. 
A valuable conclusion that can be drawn from the results is that multi-model ordinal ensembling method, which WZMedTech and SmartDSP adopted, are superior on glaucoma grading task. Specifically, 2-ordinal adopted by WZMedTech outperforms standard five fold average ensemble by a 4.52\% mean kappa improvement and 0.09\% standard deviation descent, 3-ordinal adopted by SmartDSP outperforms standard five fold average ensemble by a 7.53\% mean kappa improvement and 0.23\% standard deviation descent. This improvement comes from their divide and conquer strategy, i.e., separating this triple classification task to multiple binary classification tasks, where the models that perform the best on each sub-tasks will be picked for the final ensemble. SmartDSP also classified the sample as early-glaucoma by default when all the models do not have high confidence in their prediction, a strategy often applied by clinical experts in their decision making. In clinical practice, when multiple experts give diverse opinions and are not confident, this case will be considered suspected early glaucoma for further screening. Due to the similarity of strategies, the ordinal ensemble strategy may be of use in the real-world clinical scenario.

\begin{table}[h]
\centering
\caption{Performance of proposed ensemble strategies in the GAMMA Challenge.'Stacked' denotes pseudo-label retraining strategy adopted by FATRI-AI, 'Rescale' denotes multi-scale models ensemble strategy adopted by DIAGNOS-ETS, '3-fold', '5-fold' denote averaging model predictions on 3-fold, 5-fold validation set respectively, '2-ordinal' denotes dual-model ordinal ensemble strategy adopted by WZMedTech, '3-ordinal' denotes triple-model ordinal ensemble strategy adopted by SmartDSP. The results are shown in a format of mean(\%) $\pm$ standard deviation(\%). We run each method five times to calculate mean and standard deviation.}
\resizebox{0.8\textwidth}{!}{%
\begin{tabular}{c|cccc|c}
\hline
Ensemble  & N-Acc & E-Acc & P-Acc & G-Acc & Kappa \\ \hline
Stacked   & 90.12$\pm0.51$      & 11.82$\pm1.21$    & 84.02$\pm1.54$         & 47.45$\pm0.85$        & 71.31$\pm0.58$ \\
Rescale   & 91.82$\pm0.64$      & 24.21$\pm1.71$     & 54.81$\pm1.02$        & 40.15$\pm0.94$        & 73.59$\pm0.42$ \\
3-fold-ave & 92.57$\pm0.31$      & 37.54$\pm1.25$     & 57.92$\pm1.87$        & 47.02$\pm0.61$        & 74.55$\pm0.26$\\
5-fold-ave & 90.25$\pm0.22$     & 27.72$\pm0.66$     & 71.04$\pm0.74$        & 49.56$\pm0.31$        & 75.33$\pm0.14$ \\
2-ordinal    & 96.14$\pm0.14$      & 44.02$\pm0.27$     & 54.26$\pm0.19$        & 49.06$\pm0.08$        & 79.85$\pm0.05$ \\
3-ordinal    & 98.06$\pm0.05$      & 52.06$\pm0.10$     & 66.71$\pm0.12$        & 59.23$\pm0.05$        & 82.80$\pm0.03$ \\ \hline
\end{tabular}
\label{tab:ensemble}
}
\end{table}

\subsubsection{Effects of auxiliary tasks}
Participants are also encouraged to utilize the optic disc \& cup mask and fovea location information to improve glaucoma grading. In the GAMMA Challenge, we saw that the prior knowledge of optic disc \& cup mask helped to improve the glaucoma grading performance. EyeStar and MedIPBIT both cropped optic disc regions from the fundus images in data preprocessing. EyeStar also adopted DENet to individually process the optic disc region and polar transformed optic disc region. MedICAL utilized the optic disc \& cup mask to enhance the fundus inputs. This is also in line with the previous studies (\cite{wu2020leveraging, zhao2019weakly}) and what we found in Section \ref{sec:baseline}. In Table \ref{table:before}, we can also see that optic disc region improves the dual-branch model more than the single fundus-branch. This is because OCT volume corresponds to the optic disc region of the fundus image. Cropping the optic disc region helps to align the features of the two branches. However, we also noted that this improvement decreases in the high-performance models. As the multi-modality results shown in Table \ref{table:before}, on the models with no ordinal regression, disc region cropping improves a 6.79\% from 70.2\% to 77.0\%. However, on the models with ordinal regression, disc region cropping only improves a 4.40\% from 76.8\% to 81.2\%. We conjecture that the stronger models can extract the optic disc region on their own and do not need this prior knowledge anymore. 
%¥Few teams utilized fovea location information to improve the glaucoma grading performance. Only HZL utilized fovea location to jointly learned all three tasks by a single UNet. Clinically, fovea location is also Macular degeneration

\section{Discussion}
% teams are ranked by the performance on the enclosed test set to ensure the generalization ability of the models. However, this final ranking was affected by how the models were picked for the final submission and evaluation. We observed that the models that performed similarly on the validation set can have a gap of 2$\sim$3 \% on the test set. This disturbance will be mitigated to an end of 0.3$\sim$1 \% depending on different ensemble strategies\HBc{Please rephrase, not sure what you mean here}. It indicates that teams with a gap of less than 1\% can be regarded as equivalent. 

\subsection{Multimodal Fusion strategies in GAMMA}
In the GAMMA Challenge, we note that most multimodal fusion methods that gained high performance in GAMMA are very straightforward. Many advanced multimodal fusion techniques proposed recently were not adopted for this task. The main reason is that the fusion of fundus image and OCT volume is very different from the other more common multimodal fusion tasks. Advanced multimodal fusion algorithms can be divided into two categories: pixel-level and feature-level. Pixel-level fusion operates directly on the raw pixels of the images, making it a simple and widely used technique in medical image classification. However, it can only be applied to data with the same dimensions, such as brain magnetic resonance imaging (MRI) and computed tomography (CT) scans (\cite{sahu2014medical,singh2018ripplet}), Positron Emission Tomography (PET) and MRT scans (\cite{bhavana2015multi,shabanzade2017combination,lai2017medical}), or the chest PET and CT scans (\cite{liu2010pet}). This makes it inapplicable to our 2D-3D image fusion. Feature-level fusion, on the other hand, operates on features extracted from the images. Unlike the GAMMA methods, which typically perform fusion at the final embedding stage, these techniques often fuse multi-scale features with spatial attention, similarity matching (\cite{meher2019survey}) or domain adaptation (\cite{9514499,9627926,10.5555/3304415.3304514}) in applications such as lung-based Fluoro-D-Glucose PET (FDG PET) and MRI fusion (\cite{das2013neuro}) or Ultrasound and Single-Photon emission CT (SPECT) fusion  (\cite{tang2016nsct}). However, they are also difficult to apply to our case, as the two modalities (OCT and fundus) have a significant scale difference and lack strong spatial correspondence. OCT images typically focus on a small region near the fovea, while fundus images cover a large area of the fundus.
To our knowledge, few multimodal fusion techniques can be directly adopted for the fundus-OCT fusion task, and explains why the straightforward dual-branch concatenation model was the main choice in the GAMMA Challenge. This indicates that  more specific multimodal fusion algorithms are required in this field.

\subsection{Challenge strengths and limitations}
GAMMA was the first open initiative aiming to evaluate the possibility to develop automated methods for glaucoma grading from a combination of fundus images and OCT volume, mimicking the clinical operations of the ophthalmologists to some extent. Toward that end, the challenge provided to the community with the largest public available dataset of fundus photographs and OCT volume pairs to date. The unique characteristic of GAMMA provides a platform to establish more reliable and clinical-alike automated glaucoma classification methods, inspired by the clinical observation that the complementary fundus image and OCT can significantly improve the diagnostic accuracy of ophthalmologists (\cite{anton2021diagnostic}). In addition, GAMMA provided the glaucoma diagnostic labels according to the clinical diagnostic standard (normal/early-stage/progressive) with a high quality reference OD/OC masks and fovea positions.
These additional information is helpful to calibrate the glaucoma grading methods, as it was observed that training with fundus-derived labels have a negative impact on performance to detect truly diseased cases (\cite{phene2019deep}).
To our knowledge, GAMMA is also the only available dataset to establish valid deep neural networks for now. The only other dataset similar to GAMMA (\cite{raja2020data}) contains only 50 pairs of fundus-OCT scans, which is commonly not enough to train and evaluate the deep learning methods.

In the GAMMA challenge, the evaluation framework we designed matched the principles for evaluating retinal image analysis algorithms proposed by \cite{trucco2013validating}. Specifically, the GAMMA dataset can be easily accessed through the website associated with the Grand Challenge. Moreover, an open and uniform evaluation interface is provided on the website to automatically evaluate any results submitted. The evaluation process is exactly the same as the GAMMA Challenge. Such an online evaluation provides the further participants a platform to test their algorithms and allows them to fairly compare with the algorithms on the GAMMA leaderboard. In this way, the effectiveness of the proposed techniques can be conveniently and fairly verified, which encourages the development of the further novel algorithms.

To prevent the submitted method to be overfitting, only the training dataset is released to the registered teams in the challenge. Moreover, each team was allowed to request a maximum five times evaluation on the preliminary data set per day to adjust their algorithms. In the final stage of the challenge, the submitted methods will only be evaluated once by the test dataset as the final results. Our conclusion and analysis can therefore remain unbiased by this issue. Since the online evaluation on the preliminary dataset is limited, most teams split the released training dataset to several parts offline for the private training and evaluation. The future challenges might perhaps consider to split the dataset to four parts, for the purpose of training, validation, online evaluation and final test, respectively. Among them, the training set and the validation set are the released labeled data sets, participants can use these sets to train the models directly, or they can mix them and design their own training and validation sets or cross-validation sets. These released labeled samples will not appear in the online evaluation and final test sets.

Regarding the technical methodology, we aim to find out the most effective solution for multi-modality glaucoma grading task in GAMMA. For that purpose, different from many other challenges, we did not allow the participants to use the extra data to train their models. In addition, the source code of the wining teams is required to be submitted with their final results. These ensure the methods proposed can be fairly compared, so that the effective techniques can be identified in the challenge. We also note that many factors are tangled together to effect the final results. This often bother the readers who want to quickly find out the most effective modules on this task, like which is the most effective architecture on this task or which is the most effective ensemble strategy on this task. Thus we also do the ablation study on the wining teams to identify the effectiveness of each proposed module. Since the source code of the wining teams is submitted, we are able to correctly reproduce their methods and do the comparison. Such an ablation study provides the future researchers/developers a cookbook to design their own models.

One limitation of GAMMA is the size of the dataset. Although GAMMA is the largest fundus-OCT paired data set to date, it is still not big enough for developing capable enough deep learning models. Fundus images based glaucoma classification often provided larger datasets, for instance, our previous REFUGE (\cite{orlando2020refuge}) and REFUGE2 (\cite{fang2022refuge2}) challenges released 1600 annotated fundus images in total. Moreover, it is worth mentioning that the diverse ethnicities are lack in the GAMMA dataset, as the images correspond to a Chinese population. Although OCT may not vary too much, the fundus images of different ethnicities will be different due to changes in the pigment of the fundus. Therefore, the algorithms in the GAMMA challenge might need to be retrained before applying to a different population. These limitations should be addressed in future challenges by a large-scale multi-ethnicities collection of data, to ensure the generalization of the models. We think the main reason of the absence of explainability is we did not take explainability as a metric to rank the teams in the challenge.

In addition, we note that all the methods in the GAMMA challenge are based on black-box neural networks, and few of them are interpretable. Explainability is an important factor for the clinical adoption of CAD methods, but it is often less-explored in this field. One team, HZL, did make an effort to incorporate explainability into their model by using a multi-task learning approach that jointly learned glaucoma grading and optic disc and cup segmentation. In this way, the segmentation results could be used as evidence of the neural network's attention to the relevant parts of the image. Moreover, it is also ineffective to apply explainability methods from the deep learning community to our cases. In deep learning, explainability often refers to the ability to highlight the regions of interest in an image, such as a mustache on a face to recognize gender, or in our case, the optic disc and cup region on a fundus image to diagnose glaucoma. Many explainability techniques have been proposed for this purpose, such as the popular classification activation map (CAM) based methods. However, this type of explainability is not sufficient for medical image classification. In our case, it is not just the region but the ratio of the optic disc to the cup that is discriminative for diagnosing glaucoma. This is a problem that still needs to be studied by both machine learning and clinical research communities. In the future, we are considering adding explainability as one of the metrics in our challenges to encourage the development of these technologies for clinical applications. We have also acknowledged the limitations of the current methods and discussed the need for further research in this area in our paper.

\subsection{Clinical implications of the results and future work}
The GAMMA challenge is organized aiming to answer an open question: Should we develop automated glaucoma diagnosis based on a combination of fundus photography and OCT, like what we do clinically? Up to now, the GAMMA challenge seems to give us a preliminary but positive answer: fudus-OCT combined glaucoma grading obviously outperforms which using only fundus or OCT data. Comparing Table \ref{table:before} and Table \ref{tab:final_rank}, we can see a simple fusion of 2D fundus and 3D OCT gains about 10\% improvement against the single-modality. Another 10\% improvement can gain from the advanced design of the model. To compare with the human experts, the sensitivities (considering the classification of normal and early/progressive glaucoma) of top-3 teams (0.959, 0.918 and 0.959) have been considerably higher than reported sensitivities of junior ophthalmologists (0.694 to 0.862) (\cite{anton2021diagnostic}). 

To move a step further, could these models be applied in the real clinical scenario to automatically screen the glaucoma suspect? It is still an open question. But first, an automated, objective diagnosis system is able to to mitigate the human individual bias and to save human experts substantial time. In addition, as the non-invasive, cost-efficient and early-stage glaucoma sensitive glaucoma screening tools, fundus photographs and OCT are widely used by the clinical experts for the primary screening of glaucoma suspect (\cite{chen2019combination}). In this case, fundus and OCT combined automated glaucoma detection seems to be an appropriate solution for the large-scale community screening. These models can achieve high sensitivities (above 0.9 for the top three teams) and better overall performance than single-modality models. Although these results are limited to a specific image population, we can still envision this technique to be widely used in clinic in the future.

The functional parameters like vision field test and IOP will be considered to be contained for the automated glaucoma detection in the future. Although the tools for ONH examination, like fundus images and OCT are cost-efficient and complementary to detect early-stage glaucoma, the clinical gold standard for glaucoma is vision field, which indicates the functional impairment scale. Besides, IOP is also a valid biomarker indicating the risk of damage to the optic nerve, causing glaucoma and permanent vision loss. In the future work, we will explore the possibility of further combining IOP measurement data and visual field test data to create an automated glaucoma detection model in full accordance with the clinical glaucoma diagnosis criteria. Such models may have the chance to be deployed in both large-scale community screening scenario and in-hospital diagnosis scenario.

\section{Conclusion}
Following the clinical glaucoma screening standard, we held a challenge for automated glaucoma grading from both fundus images and OCT volumes, called Glaucoma grAding from Multi-Modality imAges (GAMMA) Challenge. In this paper, we introduced the released GAMMA dataset, the process of the challenge, the evaluation framework and the top-ranked algorithms in the challenge. Detailed comparisons and analyses are also conducted on the proposed methodologies. As the first in-depth study of fundus-OCT multi-modality glaucoma grading task, we believe GAMMA will be an essential starting point for further research on this important and clinically-relevant task.

The data and evaluation framework are publicly accessible through \url{https://gamma.grand-challenge.org/}. The code and technical reports of top-10 teams are released at \url{https://gamma.grand-challenge.org/technical-materials/}. %the  AI Studio platform at \url{https://aistudio.baidu.com/aistudio/competition/detail/119/; https://aistudio.baidu.com/aistudio/competition/detail/120/; https://aistudio.baidu.com/aistudio/competition/detail/121/}\HBc{That's too many links. Can we use only one? Btw those website are in Chinese. Do you have a link to English version?}. 
Future participants are welcome to use our dataset and submit their results on the website to benchmark their methods.

\section*{Acknowledgments}
This research was supported by the High-level Hospital Construction Project, Zhongshan Ophthalmic Center, Sun Yat-sen University (303020104).  H.~Fu was supported by AME Programmatic Fund (A20H4b0141).

%%Harvard
\bibliographystyle{model2-names.bst}\biboptions{authoryear}
\bibliography{refs}

\begin{thebibliography}{54}
\expandafter\ifx\csname natexlab\endcsname\relax\def\natexlab#1{#1}\fi
\providecommand{\url}[1]{\texttt{#1}}
\providecommand{\href}[2]{#2}
\providecommand{\path}[1]{#1}
\providecommand{\DOIprefix}{doi:}
\providecommand{\ArXivprefix}{arXiv:}
\providecommand{\URLprefix}{URL: }
\providecommand{\Pubmedprefix}{pmid:}
\providecommand{\doi}[1]{\href{http://dx.doi.org/#1}{\path{#1}}}
\providecommand{\Pubmed}[1]{\href{pmid:#1}{\path{#1}}}
\providecommand{\bibinfo}[2]{#2}
\ifx\xfnm\relax \def\xfnm[#1]{\unskip,\space#1}\fi
%Type = Article
\bibitem[{Anton et~al.(2021)Anton, Nolivos, Pazos, Fatti, Ayala,
  Mart{\'\i}nez-Prats, Peral, Poposki, Tsiroukis, Morilla-Grasa
  et~al.}]{anton2021diagnostic}
\bibinfo{author}{Anton, A.}, \bibinfo{author}{Nolivos, K.},
  \bibinfo{author}{Pazos, M.}, \bibinfo{author}{Fatti, G.},
  \bibinfo{author}{Ayala, M.E.}, \bibinfo{author}{Mart{\'\i}nez-Prats, E.},
  \bibinfo{author}{Peral, O.}, \bibinfo{author}{Poposki, V.},
  \bibinfo{author}{Tsiroukis, E.}, \bibinfo{author}{Morilla-Grasa, A.}, et~al.,
  \bibinfo{year}{2021}.
\newblock \bibinfo{title}{Diagnostic accuracy and detection rate of glaucoma
  screening with optic disk photos, optical coherence tomography images, and
  telemedicine}.
\newblock \bibinfo{journal}{Journal of Clinical Medicine} \bibinfo{volume}{11},
  \bibinfo{pages}{216}.
%Type = Article
\bibitem[{Bhavana and Krishnappa(2015)}]{bhavana2015multi}
\bibinfo{author}{Bhavana, V.}, \bibinfo{author}{Krishnappa, H.},
  \bibinfo{year}{2015}.
\newblock \bibinfo{title}{Multi-modality medical image fusion using discrete
  wavelet transform}.
\newblock \bibinfo{journal}{Procedia Computer Science} \bibinfo{volume}{70},
  \bibinfo{pages}{625--631}.
%Type = Article
\bibitem[{Bian et~al.(2021)Bian, Yuan, Ma, Yu, Wei and Zheng}]{9627926}
\bibinfo{author}{Bian, C.}, \bibinfo{author}{Yuan, C.}, \bibinfo{author}{Ma,
  K.}, \bibinfo{author}{Yu, S.}, \bibinfo{author}{Wei, D.},
  \bibinfo{author}{Zheng, Y.}, \bibinfo{year}{2021}.
\newblock \bibinfo{title}{Domain adaptation meets zero-shot learning: An
  annotation-efficient approach to multi-modality medical image segmentation}.
\newblock \bibinfo{journal}{IEEE Transactions on Medical Imaging} ,
  \bibinfo{pages}{1--1}\DOIprefix\doi{10.1109/TMI.2021.3131245}.
%Type = Inproceedings
\bibitem[{Cai et~al.(2022)Cai, Lin, He and Tang}]{cai2022corolla}
\bibinfo{author}{Cai, Z.}, \bibinfo{author}{Lin, L.}, \bibinfo{author}{He, H.},
  \bibinfo{author}{Tang, X.}, \bibinfo{year}{2022}.
\newblock \bibinfo{title}{Corolla: An efficient multi-modality fusion framework
  with supervised contrastive learning for glaucoma grading}, in:
  \bibinfo{booktitle}{2022 IEEE 19th International Symposium on Biomedical
  Imaging (ISBI)}, \bibinfo{organization}{IEEE}. pp. \bibinfo{pages}{1--4}.
%Type = Article
\bibitem[{Chen et~al.(2020)Chen, Zhang, Li, Li, Zhang, Qi, Sun and
  Jia}]{chen2020dynamic}
\bibinfo{author}{Chen, Y.}, \bibinfo{author}{Zhang, P.}, \bibinfo{author}{Li,
  Z.}, \bibinfo{author}{Li, Y.}, \bibinfo{author}{Zhang, X.},
  \bibinfo{author}{Qi, L.}, \bibinfo{author}{Sun, J.}, \bibinfo{author}{Jia,
  J.}, \bibinfo{year}{2020}.
\newblock \bibinfo{title}{Dynamic scale training for object detection}.
\newblock \bibinfo{journal}{arXiv preprint arXiv:2004.12432} .
%Type = Article
\bibitem[{Chen et~al.(2019)Chen, Zheng, Shen, Zeng, Liu and
  Li}]{chen2019combination}
\bibinfo{author}{Chen, Z.}, \bibinfo{author}{Zheng, X.}, \bibinfo{author}{Shen,
  H.}, \bibinfo{author}{Zeng, Z.}, \bibinfo{author}{Liu, Q.},
  \bibinfo{author}{Li, Z.}, \bibinfo{year}{2019}.
\newblock \bibinfo{title}{Combination of enhanced depth imaging optical
  coherence tomography and fundus images for glaucoma screening}.
\newblock \bibinfo{journal}{Journal of Medical Systems} \bibinfo{volume}{43},
  \bibinfo{pages}{1--12}.
%Type = Article
\bibitem[{Das and Kundu(2013)}]{das2013neuro}
\bibinfo{author}{Das, S.}, \bibinfo{author}{Kundu, M.K.}, \bibinfo{year}{2013}.
\newblock \bibinfo{title}{A neuro-fuzzy approach for medical image fusion}.
\newblock \bibinfo{journal}{IEEE transactions on biomedical engineering}
  \bibinfo{volume}{60}, \bibinfo{pages}{3347--3353}.
%Type = Inproceedings
\bibitem[{Dou et~al.(2018)Dou, Ouyang, Chen, Chen and
  Heng}]{10.5555/3304415.3304514}
\bibinfo{author}{Dou, Q.}, \bibinfo{author}{Ouyang, C.}, \bibinfo{author}{Chen,
  C.}, \bibinfo{author}{Chen, H.}, \bibinfo{author}{Heng, P.A.},
  \bibinfo{year}{2018}.
\newblock \bibinfo{title}{Unsupervised cross-modality domain adaptation of
  convnets for biomedical image segmentations with adversarial loss}, in:
  \bibinfo{booktitle}{Proceedings of the 27th International Joint Conference on
  Artificial Intelligence}, \bibinfo{publisher}{AAAI Press}. p.
  \bibinfo{pages}{691–697}.
%Type = Article
\bibitem[{Fang et~al.(2022)Fang, Li, Fu, Sun, Cao, Son, Yu, Zhang, Yuan, Bian
  et~al.}]{fang2022refuge2}
\bibinfo{author}{Fang, H.}, \bibinfo{author}{Li, F.}, \bibinfo{author}{Fu, H.},
  \bibinfo{author}{Sun, X.}, \bibinfo{author}{Cao, X.}, \bibinfo{author}{Son,
  J.}, \bibinfo{author}{Yu, S.}, \bibinfo{author}{Zhang, M.},
  \bibinfo{author}{Yuan, C.}, \bibinfo{author}{Bian, C.}, et~al.,
  \bibinfo{year}{2022}.
\newblock \bibinfo{title}{Refuge2 challenge: Treasure for multi-domain learning
  in glaucoma assessment}.
\newblock \bibinfo{journal}{arXiv preprint arXiv:2202.08994} .
%Type = Inproceedings
\bibitem[{Fang et~al.(2021)Fang, Shang, Fu, Li, Zhang and Xu}]{fang2021multi}
\bibinfo{author}{Fang, H.}, \bibinfo{author}{Shang, F.}, \bibinfo{author}{Fu,
  H.}, \bibinfo{author}{Li, F.}, \bibinfo{author}{Zhang, X.},
  \bibinfo{author}{Xu, Y.}, \bibinfo{year}{2021}.
\newblock \bibinfo{title}{Multi-modality images analysis: A baseline for
  glaucoma grading via deep learning}, in: \bibinfo{booktitle}{International
  Workshop on Ophthalmic Medical Image Analysis},
  \bibinfo{organization}{Springer}. pp. \bibinfo{pages}{139--147}.
%Type = Article
\bibitem[{Fu et~al.(2018a)Fu, Cheng, Xu, Wong, Liu and Cao}]{fu2018joint}
\bibinfo{author}{Fu, H.}, \bibinfo{author}{Cheng, J.}, \bibinfo{author}{Xu,
  Y.}, \bibinfo{author}{Wong, D.W.K.}, \bibinfo{author}{Liu, J.},
  \bibinfo{author}{Cao, X.}, \bibinfo{year}{2018}a.
\newblock \bibinfo{title}{Joint optic disc and cup segmentation based on
  multi-label deep network and polar transformation}.
\newblock \bibinfo{journal}{IEEE transactions on medical imaging}
  \bibinfo{volume}{37}, \bibinfo{pages}{1597--1605}.
%Type = Article
\bibitem[{Fu et~al.(2018b)Fu, Cheng et~al.}]{Fu2018DiscAware}
\bibinfo{author}{Fu, H.}, \bibinfo{author}{Cheng, J.}, et~al.,
  \bibinfo{year}{2018}b.
\newblock \bibinfo{title}{{Disc-Aware Ensemble Network for Glaucoma Screening
  From Fundus Image}}.
\newblock \bibinfo{journal}{IEEE Transactions on Medical Imaging}
  \bibinfo{volume}{37}, \bibinfo{pages}{2493--2501}.
%Type = Article
\bibitem[{Fu et~al.(2017)Fu, Xu, Lin, Zhang, Wong, Liu, Frangi, Baskaran and
  Aung}]{fu2017segmentation}
\bibinfo{author}{Fu, H.}, \bibinfo{author}{Xu, Y.}, \bibinfo{author}{Lin, S.},
  \bibinfo{author}{Zhang, X.}, \bibinfo{author}{Wong, D.W.K.},
  \bibinfo{author}{Liu, J.}, \bibinfo{author}{Frangi, A.F.},
  \bibinfo{author}{Baskaran, M.}, \bibinfo{author}{Aung, T.},
  \bibinfo{year}{2017}.
\newblock \bibinfo{title}{Segmentation and quantification for angle-closure
  glaucoma assessment in anterior segment oct}.
\newblock \bibinfo{journal}{IEEE transactions on medical imaging}
  \bibinfo{volume}{36}, \bibinfo{pages}{1930--1938}.
%Type = Article
\bibitem[{Ge et~al.(2022)Ge, Pereira, Mitteregger, Berry, Zhang, Hausmann,
  Zhang, Schintlmeister, Wagner and Cheng}]{ge2022srs}
\bibinfo{author}{Ge, X.}, \bibinfo{author}{Pereira, F.C.},
  \bibinfo{author}{Mitteregger, M.}, \bibinfo{author}{Berry, D.},
  \bibinfo{author}{Zhang, M.}, \bibinfo{author}{Hausmann, B.},
  \bibinfo{author}{Zhang, J.}, \bibinfo{author}{Schintlmeister, A.},
  \bibinfo{author}{Wagner, M.}, \bibinfo{author}{Cheng, J.X.},
  \bibinfo{year}{2022}.
\newblock \bibinfo{title}{Srs-fish: A high-throughput platform linking
  microbiome metabolism to identity at the single-cell level}.
\newblock \bibinfo{journal}{Proceedings of the National Academy of Sciences}
  \bibinfo{volume}{119}, \bibinfo{pages}{e2203519119}.
%Type = Article
\bibitem[{Han et~al.(2022)Han, Qi, Yu, Zhou, Zheng, Shi and Gao}]{9514499}
\bibinfo{author}{Han, X.}, \bibinfo{author}{Qi, L.}, \bibinfo{author}{Yu, Q.},
  \bibinfo{author}{Zhou, Z.}, \bibinfo{author}{Zheng, Y.},
  \bibinfo{author}{Shi, Y.}, \bibinfo{author}{Gao, Y.}, \bibinfo{year}{2022}.
\newblock \bibinfo{title}{Deep symmetric adaptation network for cross-modality
  medical image segmentation}.
\newblock \bibinfo{journal}{IEEE Transactions on Medical Imaging}
  \bibinfo{volume}{41}, \bibinfo{pages}{121--132}.
\newblock \DOIprefix\doi{10.1109/TMI.2021.3105046}.
%Type = Article
\bibitem[{Hancox~OD(1999)}]{hancox1999optic}
\bibinfo{author}{Hancox~OD, M.D.}, \bibinfo{year}{1999}.
\newblock \bibinfo{title}{Optic disc size, an important consideration in the
  glaucoma evaluation}.
\newblock \bibinfo{journal}{Clinical Eye and Vision Care} \bibinfo{volume}{11},
  \bibinfo{pages}{59--62}.
%Type = Article
\bibitem[{He et~al.(2022)He, Lin, Cai and Tang}]{he2022joined}
\bibinfo{author}{He, H.}, \bibinfo{author}{Lin, L.}, \bibinfo{author}{Cai, Z.},
  \bibinfo{author}{Tang, X.}, \bibinfo{year}{2022}.
\newblock \bibinfo{title}{Joined: Prior guided multi-task learning for joint
  optic disc/cup segmentation and fovea detection}.
\newblock \bibinfo{journal}{arXiv preprint arXiv:2203.00461} .
%Type = Inproceedings
\bibitem[{He et~al.(2016)He, Zhang, Ren and Sun}]{he2016deep}
\bibinfo{author}{He, K.}, \bibinfo{author}{Zhang, X.}, \bibinfo{author}{Ren,
  S.}, \bibinfo{author}{Sun, J.}, \bibinfo{year}{2016}.
\newblock \bibinfo{title}{Deep residual learning for image recognition}, in:
  \bibinfo{booktitle}{Proceedings of the IEEE conference on computer vision and
  pattern recognition}, pp. \bibinfo{pages}{770--778}.
%Type = Inproceedings
\bibitem[{Huang et~al.(2021)Huang, Lu, Cheng and He}]{huang2021fapn}
\bibinfo{author}{Huang, S.}, \bibinfo{author}{Lu, Z.}, \bibinfo{author}{Cheng,
  R.}, \bibinfo{author}{He, C.}, \bibinfo{year}{2021}.
\newblock \bibinfo{title}{Fapn: Feature-aligned pyramid network for dense image
  prediction}, in: \bibinfo{booktitle}{Proceedings of the IEEE/CVF
  International Conference on Computer Vision}, pp. \bibinfo{pages}{864--873}.
%Type = Article
\bibitem[{Jonas et~al.(2000)Jonas, Bergua, Schmitz-Valckenberg,
  Papastathopoulos and Budde}]{jonas2000ranking}
\bibinfo{author}{Jonas, J.B.}, \bibinfo{author}{Bergua, A.},
  \bibinfo{author}{Schmitz-Valckenberg, P.}, \bibinfo{author}{Papastathopoulos,
  K.I.}, \bibinfo{author}{Budde, W.M.}, \bibinfo{year}{2000}.
\newblock \bibinfo{title}{Ranking of optic disc variables for detection of
  glaucomatous optic nerve damage}.
\newblock \bibinfo{journal}{Investigative Ophthalmology \& Visual Science}
  \bibinfo{volume}{41}, \bibinfo{pages}{1764--1773}.
%Type = Article
\bibitem[{Jonas et~al.(1999)Jonas, Budde and
  Panda-Jonas}]{jonas1999ophthalmoscopic}
\bibinfo{author}{Jonas, J.B.}, \bibinfo{author}{Budde, W.M.},
  \bibinfo{author}{Panda-Jonas, S.}, \bibinfo{year}{1999}.
\newblock \bibinfo{title}{Ophthalmoscopic evaluation of the optic nerve head}.
\newblock \bibinfo{journal}{Survey of ophthalmology} \bibinfo{volume}{43},
  \bibinfo{pages}{293--320}.
%Type = Article
\bibitem[{Kingma and Ba(2014)}]{kingma2014adam}
\bibinfo{author}{Kingma, D.P.}, \bibinfo{author}{Ba, J.}, \bibinfo{year}{2014}.
\newblock \bibinfo{title}{Adam: A method for stochastic optimization}.
\newblock \bibinfo{journal}{arXiv preprint arXiv:1412.6980} .
%Type = Article
\bibitem[{Lai et~al.(2017)Lai, Wang, He and Borjer}]{lai2017medical}
\bibinfo{author}{Lai, S.}, \bibinfo{author}{Wang, J.}, \bibinfo{author}{He,
  C.}, \bibinfo{author}{Borjer, T.H.}, \bibinfo{year}{2017}.
\newblock \bibinfo{title}{Medical image fusion combined with accelerated
  non-negative matrix factorization and expanded laplacian energy in shearlet
  domain.}
\newblock \bibinfo{journal}{Journal of Engineering Science \& Technology
  Review} \bibinfo{volume}{10}.
%Type = Article
\bibitem[{Li et~al.(2020)Li, Song, Chen, Xiong, Li, Zhong, Tang, Fan, Lam, Pan
  et~al.}]{li2020development}
\bibinfo{author}{Li, F.}, \bibinfo{author}{Song, D.}, \bibinfo{author}{Chen,
  H.}, \bibinfo{author}{Xiong, J.}, \bibinfo{author}{Li, X.},
  \bibinfo{author}{Zhong, H.}, \bibinfo{author}{Tang, G.},
  \bibinfo{author}{Fan, S.}, \bibinfo{author}{Lam, D.S.}, \bibinfo{author}{Pan,
  W.}, et~al., \bibinfo{year}{2020}.
\newblock \bibinfo{title}{Development and clinical deployment of a
  smartphone-based visual field deep learning system for glaucoma detection}.
\newblock \bibinfo{journal}{NPJ digital medicine} \bibinfo{volume}{3},
  \bibinfo{pages}{1--8}.
%Type = Article
\bibitem[{Li et~al.(2022)Li, Li, Kou, Yang, Hu, Peng and Li}]{li2022dynamic}
\bibinfo{author}{Li, L.}, \bibinfo{author}{Li, H.}, \bibinfo{author}{Kou, G.},
  \bibinfo{author}{Yang, D.}, \bibinfo{author}{Hu, W.}, \bibinfo{author}{Peng,
  J.}, \bibinfo{author}{Li, S.}, \bibinfo{year}{2022}.
\newblock \bibinfo{title}{Dynamic camouflage characteristics of a thermal
  infrared film inspired by honeycomb structure}.
\newblock \bibinfo{journal}{Journal of Bionic Engineering}
  \bibinfo{volume}{19}, \bibinfo{pages}{458--470}.
%Type = Article
\bibitem[{Li et~al.(2021)Li, Sui, Luo, Xu, Liu and Goh}]{li2021medical}
\bibinfo{author}{Li, S.}, \bibinfo{author}{Sui, X.}, \bibinfo{author}{Luo, X.},
  \bibinfo{author}{Xu, X.}, \bibinfo{author}{Liu, Y.}, \bibinfo{author}{Goh,
  R.S.M.}, \bibinfo{year}{2021}.
\newblock \bibinfo{title}{Medical image segmentation using
  squeeze-and-expansion transformers}.
\newblock \bibinfo{journal}{arXiv preprint arXiv:2105.09511} .
%Type = Inproceedings
\bibitem[{Liu et~al.(2010)Liu, Yang and Sun}]{liu2010pet}
\bibinfo{author}{Liu, Y.}, \bibinfo{author}{Yang, J.}, \bibinfo{author}{Sun,
  J.}, \bibinfo{year}{2010}.
\newblock \bibinfo{title}{Pet/ct medical image fusion algorithm based on
  multiwavelet transform}, in: \bibinfo{booktitle}{2010 2nd International
  Conference on Advanced Computer Control}, \bibinfo{organization}{IEEE}. pp.
  \bibinfo{pages}{264--268}.
%Type = Inproceedings
\bibitem[{Liu et~al.(2021)Liu, Lin, Cao, Hu, Wei, Zhang, Lin and
  Guo}]{liu2021swin}
\bibinfo{author}{Liu, Z.}, \bibinfo{author}{Lin, Y.}, \bibinfo{author}{Cao,
  Y.}, \bibinfo{author}{Hu, H.}, \bibinfo{author}{Wei, Y.},
  \bibinfo{author}{Zhang, Z.}, \bibinfo{author}{Lin, S.}, \bibinfo{author}{Guo,
  B.}, \bibinfo{year}{2021}.
\newblock \bibinfo{title}{Swin transformer: Hierarchical vision transformer
  using shifted windows}, in: \bibinfo{booktitle}{Proceedings of the IEEE/CVF
  International Conference on Computer Vision}, pp.
  \bibinfo{pages}{10012--10022}.
%Type = Article
\bibitem[{Meher et~al.(2019)Meher, Agrawal, Panda and
  Abraham}]{meher2019survey}
\bibinfo{author}{Meher, B.}, \bibinfo{author}{Agrawal, S.},
  \bibinfo{author}{Panda, R.}, \bibinfo{author}{Abraham, A.},
  \bibinfo{year}{2019}.
\newblock \bibinfo{title}{A survey on region based image fusion methods}.
\newblock \bibinfo{journal}{Information Fusion} \bibinfo{volume}{48},
  \bibinfo{pages}{119--132}.
%Type = Article
\bibitem[{Morgan et~al.(2005)Morgan, Sheen, North, Choong and
  Ansari}]{morgan2005digital}
\bibinfo{author}{Morgan, J.E.}, \bibinfo{author}{Sheen, N.J.L.},
  \bibinfo{author}{North, R.V.}, \bibinfo{author}{Choong, Y.},
  \bibinfo{author}{Ansari, E.}, \bibinfo{year}{2005}.
\newblock \bibinfo{title}{Digital imaging of the optic nerve head: monoscopic
  and stereoscopic analysis}.
\newblock \bibinfo{journal}{British journal of ophthalmology}
  \bibinfo{volume}{89}, \bibinfo{pages}{879--884}.
%Type = Article
\bibitem[{Nayak et~al.(2009)Nayak, Acharya, Bhat, Shetty and
  Lim}]{nayak2009automated}
\bibinfo{author}{Nayak, J.}, \bibinfo{author}{Acharya, R.},
  \bibinfo{author}{Bhat, P.S.}, \bibinfo{author}{Shetty, N.},
  \bibinfo{author}{Lim, T.C.}, \bibinfo{year}{2009}.
\newblock \bibinfo{title}{Automated diagnosis of glaucoma using digital fundus
  images}.
\newblock \bibinfo{journal}{Journal of medical systems} \bibinfo{volume}{33},
  \bibinfo{pages}{337--346}.
%Type = Inproceedings
\bibitem[{Niu et~al.(2016)Niu, Zhou, Wang, Gao and Hua}]{niu2016ordinal}
\bibinfo{author}{Niu, Z.}, \bibinfo{author}{Zhou, M.}, \bibinfo{author}{Wang,
  L.}, \bibinfo{author}{Gao, X.}, \bibinfo{author}{Hua, G.},
  \bibinfo{year}{2016}.
\newblock \bibinfo{title}{Ordinal regression with multiple output cnn for age
  estimation}, in: \bibinfo{booktitle}{Proceedings of the IEEE conference on
  computer vision and pattern recognition}, pp. \bibinfo{pages}{4920--4928}.
%Type = Article
\bibitem[{Orlando et~al.(2020)Orlando, Fu, Breda, van Keer, Bathula,
  Diaz-Pinto, Fang, Heng, Kim, Lee et~al.}]{orlando2020refuge}
\bibinfo{author}{Orlando, J.I.}, \bibinfo{author}{Fu, H.},
  \bibinfo{author}{Breda, J.B.}, \bibinfo{author}{van Keer, K.},
  \bibinfo{author}{Bathula, D.R.}, \bibinfo{author}{Diaz-Pinto, A.},
  \bibinfo{author}{Fang, R.}, \bibinfo{author}{Heng, P.A.},
  \bibinfo{author}{Kim, J.}, \bibinfo{author}{Lee, J.}, et~al.,
  \bibinfo{year}{2020}.
\newblock \bibinfo{title}{Refuge challenge: A unified framework for evaluating
  automated methods for glaucoma assessment from fundus photographs}.
\newblock \bibinfo{journal}{Medical image analysis} \bibinfo{volume}{59},
  \bibinfo{pages}{101570}.
%Type = Article
\bibitem[{Phene et~al.(2019)Phene, Carter~Dunn, Hammel, Liu, Krause, Kitade
  et~al.}]{phene2019deep}
\bibinfo{author}{Phene, S.}, \bibinfo{author}{Carter~Dunn, R.},
  \bibinfo{author}{Hammel, N.}, \bibinfo{author}{Liu, Y.},
  \bibinfo{author}{Krause, J.}, \bibinfo{author}{Kitade, N.}, et~al.,
  \bibinfo{year}{2019}.
\newblock \bibinfo{title}{Deep learning to assess glaucoma risk and associated
  features in fundus images}.
\newblock \bibinfo{journal}{arXiv preprint arXiv:1812.08911} .
%Type = Article
\bibitem[{Raja et~al.(2020)Raja, Akram, Khawaja, Arslan, Ramzan and
  Nazir}]{raja2020data}
\bibinfo{author}{Raja, H.}, \bibinfo{author}{Akram, M.U.},
  \bibinfo{author}{Khawaja, S.G.}, \bibinfo{author}{Arslan, M.},
  \bibinfo{author}{Ramzan, A.}, \bibinfo{author}{Nazir, N.},
  \bibinfo{year}{2020}.
\newblock \bibinfo{title}{Data on oct and fundus images for the detection of
  glaucoma}.
\newblock \bibinfo{journal}{Data in brief} \bibinfo{volume}{29},
  \bibinfo{pages}{105342}.
%Type = Inproceedings
\bibitem[{Redmon et~al.(2016)Redmon, Divvala, Girshick and
  Farhadi}]{redmon2016you}
\bibinfo{author}{Redmon, J.}, \bibinfo{author}{Divvala, S.},
  \bibinfo{author}{Girshick, R.}, \bibinfo{author}{Farhadi, A.},
  \bibinfo{year}{2016}.
\newblock \bibinfo{title}{You only look once: Unified, real-time object
  detection}, in: \bibinfo{booktitle}{Proceedings of the IEEE conference on
  computer vision and pattern recognition}, pp. \bibinfo{pages}{779--788}.
%Type = Article
\bibitem[{Resnikoff et~al.(2004)Resnikoff, Pascolini, Etya'Ale, Kocur,
  Pararajasegaram, Pokharel and Mariotti}]{resnikoff2004global}
\bibinfo{author}{Resnikoff, S.}, \bibinfo{author}{Pascolini, D.},
  \bibinfo{author}{Etya'Ale, D.}, \bibinfo{author}{Kocur, I.},
  \bibinfo{author}{Pararajasegaram, R.}, \bibinfo{author}{Pokharel, G.P.},
  \bibinfo{author}{Mariotti, S.P.}, \bibinfo{year}{2004}.
\newblock \bibinfo{title}{Global data on visual impairment in the year 2002}.
\newblock \bibinfo{journal}{Bulletin of the world health organization}
  \bibinfo{volume}{82}, \bibinfo{pages}{844--851}.
%Type = Inproceedings
\bibitem[{Ronneberger et~al.(2015)Ronneberger, Fischer and
  Brox}]{ronneberger2015u}
\bibinfo{author}{Ronneberger, O.}, \bibinfo{author}{Fischer, P.},
  \bibinfo{author}{Brox, T.}, \bibinfo{year}{2015}.
\newblock \bibinfo{title}{U-net: Convolutional networks for biomedical image
  segmentation}, in: \bibinfo{booktitle}{International Conference on Medical
  image computing and computer-assisted intervention},
  \bibinfo{organization}{Springer}. pp. \bibinfo{pages}{234--241}.
%Type = Inproceedings
\bibitem[{Rosenthal et~al.(2016)Rosenthal, Ritter, Kowerko and
  Heine}]{rosenthal2016ophthalvis}
\bibinfo{author}{Rosenthal, P.}, \bibinfo{author}{Ritter, M.},
  \bibinfo{author}{Kowerko, D.}, \bibinfo{author}{Heine, C.},
  \bibinfo{year}{2016}.
\newblock \bibinfo{title}{Ophthalvis-making data analytics of optical coherence
  tomography reproducible.}, in: \bibinfo{booktitle}{EuroRV$^3$@ EuroVis}, pp.
  \bibinfo{pages}{9--13}.
%Type = Inproceedings
\bibitem[{Sahu et~al.(2014)Sahu, Bhateja, Krishn et~al.}]{sahu2014medical}
\bibinfo{author}{Sahu, A.}, \bibinfo{author}{Bhateja, V.},
  \bibinfo{author}{Krishn, A.}, et~al., \bibinfo{year}{2014}.
\newblock \bibinfo{title}{Medical image fusion with laplacian pyramids}, in:
  \bibinfo{booktitle}{2014 International conference on medical imaging,
  m-health and emerging communication systems (MedCom)},
  \bibinfo{organization}{IEEE}. pp. \bibinfo{pages}{448--453}.
%Type = Inproceedings
\bibitem[{Shabanzade and Ghassemian(2017)}]{shabanzade2017combination}
\bibinfo{author}{Shabanzade, F.}, \bibinfo{author}{Ghassemian, H.},
  \bibinfo{year}{2017}.
\newblock \bibinfo{title}{Combination of wavelet and contourlet transforms for
  pet and mri image fusion}, in: \bibinfo{booktitle}{2017 artificial
  intelligence and signal processing conference (AISP)},
  \bibinfo{organization}{IEEE}. pp. \bibinfo{pages}{178--183}.
%Type = Article
\bibitem[{Singh and Anand(2018)}]{singh2018ripplet}
\bibinfo{author}{Singh, S.}, \bibinfo{author}{Anand, R.S.},
  \bibinfo{year}{2018}.
\newblock \bibinfo{title}{Ripplet domain fusion approach for ct and mr medical
  image information}.
\newblock \bibinfo{journal}{Biomedical Signal Processing and Control}
  \bibinfo{volume}{46}, \bibinfo{pages}{281--292}.
%Type = Inproceedings
\bibitem[{Tan and Le(2019)}]{tan2019efficientnet}
\bibinfo{author}{Tan, M.}, \bibinfo{author}{Le, Q.}, \bibinfo{year}{2019}.
\newblock \bibinfo{title}{Efficientnet: Rethinking model scaling for
  convolutional neural networks}, in: \bibinfo{booktitle}{International
  conference on machine learning}, \bibinfo{organization}{PMLR}. pp.
  \bibinfo{pages}{6105--6114}.
%Type = Article
\bibitem[{Tang et~al.(2016)Tang, Li, Qian, Zhang and Pan}]{tang2016nsct}
\bibinfo{author}{Tang, L.}, \bibinfo{author}{Li, L.}, \bibinfo{author}{Qian,
  J.}, \bibinfo{author}{Zhang, J.}, \bibinfo{author}{Pan, J.S.},
  \bibinfo{year}{2016}.
\newblock \bibinfo{title}{Nsct-based multimodal medical image fusion with
  sparse representation and pulse coupled neural network.}
\newblock \bibinfo{journal}{J. Inf. Hiding Multim. Signal Process.}
  \bibinfo{volume}{7}, \bibinfo{pages}{1306--1316}.
%Type = Inproceedings
\bibitem[{Tran et~al.(2015)Tran, Bourdev, Fergus, Torresani and
  Paluri}]{tran2015learning}
\bibinfo{author}{Tran, D.}, \bibinfo{author}{Bourdev, L.},
  \bibinfo{author}{Fergus, R.}, \bibinfo{author}{Torresani, L.},
  \bibinfo{author}{Paluri, M.}, \bibinfo{year}{2015}.
\newblock \bibinfo{title}{Learning spatiotemporal features with 3d
  convolutional networks}, in: \bibinfo{booktitle}{Proceedings of the IEEE
  international conference on computer vision}, pp.
  \bibinfo{pages}{4489--4497}.
%Type = Article
\bibitem[{Trobe et~al.(1980)Trobe, Glaser and Cassady}]{trobe1980optic}
\bibinfo{author}{Trobe, J.D.}, \bibinfo{author}{Glaser, J.S.},
  \bibinfo{author}{Cassady, J.C.}, \bibinfo{year}{1980}.
\newblock \bibinfo{title}{Optic atrophy: differential diagnosis by fundus
  observation alone}.
\newblock \bibinfo{journal}{Archives of Ophthalmology} \bibinfo{volume}{98},
  \bibinfo{pages}{1040--1045}.
%Type = Article
\bibitem[{Trucco et~al.(2013)Trucco, Ruggeri, Karnowski, Giancardo, Chaum,
  Hubschman, Al-Diri, Cheung, Wong, Abramoff et~al.}]{trucco2013validating}
\bibinfo{author}{Trucco, E.}, \bibinfo{author}{Ruggeri, A.},
  \bibinfo{author}{Karnowski, T.}, \bibinfo{author}{Giancardo, L.},
  \bibinfo{author}{Chaum, E.}, \bibinfo{author}{Hubschman, J.P.},
  \bibinfo{author}{Al-Diri, B.}, \bibinfo{author}{Cheung, C.Y.},
  \bibinfo{author}{Wong, D.}, \bibinfo{author}{Abramoff, M.}, et~al.,
  \bibinfo{year}{2013}.
\newblock \bibinfo{title}{Validating retinal fundus image analysis algorithms:
  issues and a proposal}.
\newblock \bibinfo{journal}{Investigative ophthalmology \& visual science}
  \bibinfo{volume}{54}, \bibinfo{pages}{3546--3559}.
%Type = Inproceedings
\bibitem[{Vaswani et~al.(2017)Vaswani, Shazeer, Parmar, Uszkoreit, Jones,
  Gomez, Kaiser and Polosukhin}]{vaswani2017attention}
\bibinfo{author}{Vaswani, A.}, \bibinfo{author}{Shazeer, N.},
  \bibinfo{author}{Parmar, N.}, \bibinfo{author}{Uszkoreit, J.},
  \bibinfo{author}{Jones, L.}, \bibinfo{author}{Gomez, A.N.},
  \bibinfo{author}{Kaiser, {\L}.}, \bibinfo{author}{Polosukhin, I.},
  \bibinfo{year}{2017}.
\newblock \bibinfo{title}{Attention is all you need}, in:
  \bibinfo{booktitle}{Advances in neural information processing systems}, pp.
  \bibinfo{pages}{5998--6008}.
%Type = Article
\bibitem[{Vos et~al.(2016)Vos, Allen, Arora, Barber, Bhutta, Brown, Carter,
  Casey, Charlson, Chen et~al.}]{vos2016global}
\bibinfo{author}{Vos, T.}, \bibinfo{author}{Allen, C.}, \bibinfo{author}{Arora,
  M.}, \bibinfo{author}{Barber, R.M.}, \bibinfo{author}{Bhutta, Z.A.},
  \bibinfo{author}{Brown, A.}, \bibinfo{author}{Carter, A.},
  \bibinfo{author}{Casey, D.C.}, \bibinfo{author}{Charlson, F.J.},
  \bibinfo{author}{Chen, A.Z.}, et~al., \bibinfo{year}{2016}.
\newblock \bibinfo{title}{Global, regional, and national incidence, prevalence,
  and years lived with disability for 310 diseases and injuries, 1990--2015: a
  systematic analysis for the global burden of disease study 2015}.
\newblock \bibinfo{journal}{The lancet} \bibinfo{volume}{388},
  \bibinfo{pages}{1545--1602}.
%Type = Inproceedings
\bibitem[{Wu et~al.(2020)Wu, Yu, Chen, Ma, Fu, Liu, Di and
  Zheng}]{wu2020leveraging}
\bibinfo{author}{Wu, J.}, \bibinfo{author}{Yu, S.}, \bibinfo{author}{Chen, W.},
  \bibinfo{author}{Ma, K.}, \bibinfo{author}{Fu, R.}, \bibinfo{author}{Liu,
  H.}, \bibinfo{author}{Di, X.}, \bibinfo{author}{Zheng, Y.},
  \bibinfo{year}{2020}.
\newblock \bibinfo{title}{Leveraging undiagnosed data for glaucoma
  classification with teacher-student learning}, in:
  \bibinfo{booktitle}{International Conference on Medical Image Computing and
  Computer-Assisted Intervention}, \bibinfo{organization}{Springer}. pp.
  \bibinfo{pages}{731--740}.
%Type = Article
\bibitem[{Xie et~al.(2020)Xie, Liu, Cao, Qiu, Duan, Garibaldi and
  Qiu}]{xie2020end}
\bibinfo{author}{Xie, R.}, \bibinfo{author}{Liu, J.}, \bibinfo{author}{Cao,
  R.}, \bibinfo{author}{Qiu, C.S.}, \bibinfo{author}{Duan, J.},
  \bibinfo{author}{Garibaldi, J.}, \bibinfo{author}{Qiu, G.},
  \bibinfo{year}{2020}.
\newblock \bibinfo{title}{End-to-end fovea localisation in colour fundus images
  with a hierarchical deep regression network}.
\newblock \bibinfo{journal}{IEEE Transactions on Medical Imaging}
  \bibinfo{volume}{40}, \bibinfo{pages}{116--128}.
%Type = Article
\bibitem[{Xiong et~al.(2021)Xiong, Li, Song, Tang, He, Gao, Zhang, Cheng, Song,
  Lin et~al.}]{xiong2021multimodal}
\bibinfo{author}{Xiong, J.}, \bibinfo{author}{Li, F.}, \bibinfo{author}{Song,
  D.}, \bibinfo{author}{Tang, G.}, \bibinfo{author}{He, J.},
  \bibinfo{author}{Gao, K.}, \bibinfo{author}{Zhang, H.},
  \bibinfo{author}{Cheng, W.}, \bibinfo{author}{Song, Y.},
  \bibinfo{author}{Lin, F.}, et~al., \bibinfo{year}{2021}.
\newblock \bibinfo{title}{Multimodal machine learning using visual fields and
  peripapillary circular oct scans in detection of glaucomatous optic
  neuropathy}.
\newblock \bibinfo{journal}{Ophthalmology} .
%Type = Article
\bibitem[{Zhang et~al.(2020)Zhang, Zhao, Lin, Tan and Cheng}]{zhang2020high}
\bibinfo{author}{Zhang, J.}, \bibinfo{author}{Zhao, J.}, \bibinfo{author}{Lin,
  H.}, \bibinfo{author}{Tan, Y.}, \bibinfo{author}{Cheng, J.X.},
  \bibinfo{year}{2020}.
\newblock \bibinfo{title}{High-speed chemical imaging by dense-net learning of
  femtosecond stimulated raman scattering}.
\newblock \bibinfo{journal}{The journal of physical chemistry letters}
  \bibinfo{volume}{11}, \bibinfo{pages}{8573--8578}.
%Type = Inproceedings
\bibitem[{Zhao et~al.(2019)Zhao, Liao, Zou, Chen and Li}]{zhao2019weakly}
\bibinfo{author}{Zhao, R.}, \bibinfo{author}{Liao, W.}, \bibinfo{author}{Zou,
  B.}, \bibinfo{author}{Chen, Z.}, \bibinfo{author}{Li, S.},
  \bibinfo{year}{2019}.
\newblock \bibinfo{title}{Weakly-supervised simultaneous evidence
  identification and segmentation for automated glaucoma diagnosis}, in:
  \bibinfo{booktitle}{Proceedings of the AAAI Conference on Artificial
  Intelligence}, pp. \bibinfo{pages}{809--816}.

\end{thebibliography}
\clearpage
\section *{Appendix}
In the following sections, we briefly introduce the methods proposed for the auxiliary tasks. 
%The final ranking of the teams considering all three tasks is also reported.

\begin{figure*}[h]
\centering
\includegraphics[width=1\linewidth]{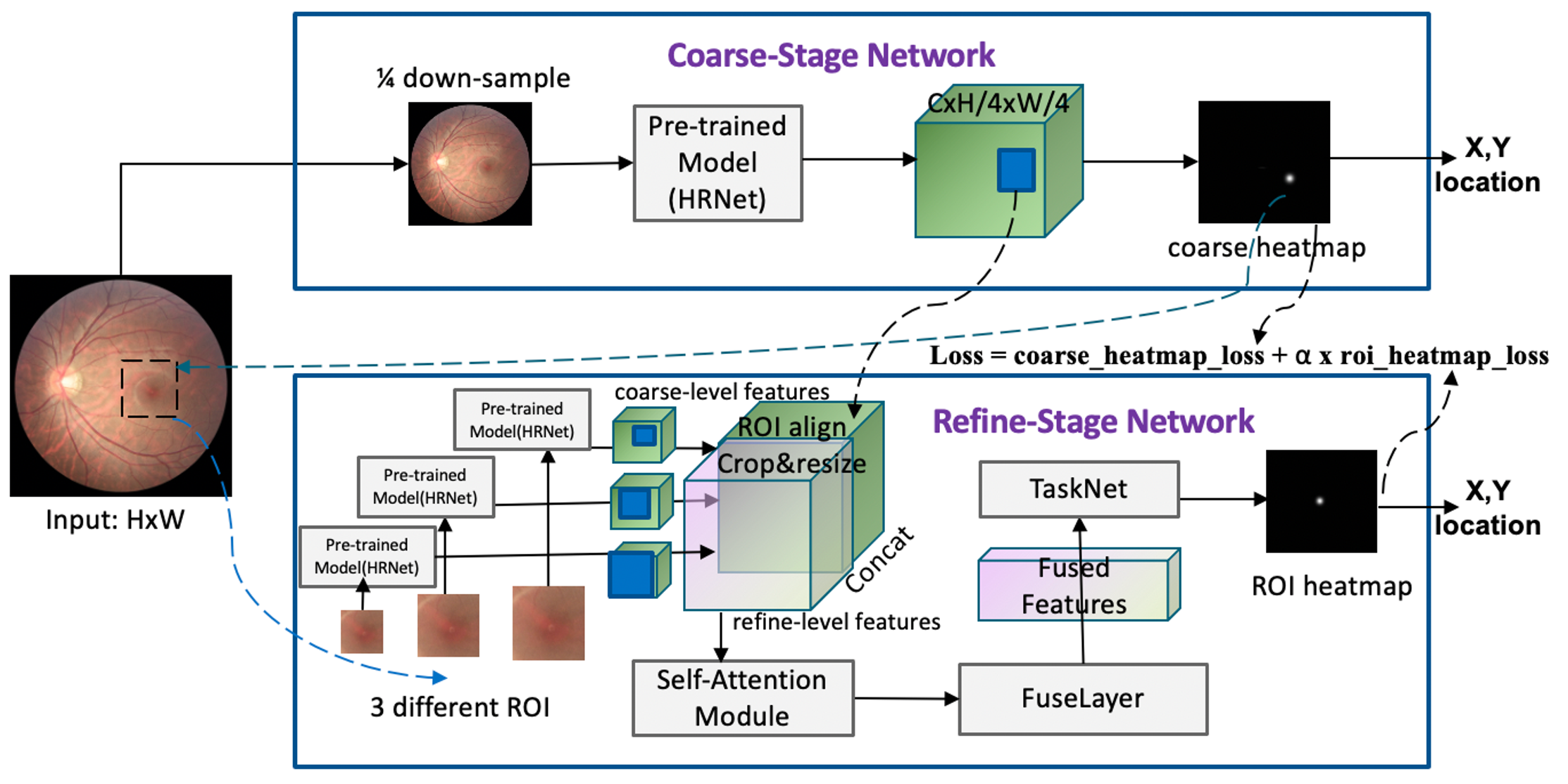}
\caption{An illustration of TSSAA proposed by EyeStar for fovea localization. TSSAA first predicts a coarse heatmap in the coarse stage. Then multi-scale ROI is cropped from the raw image as the input of the subsequent refine stage. In the refine stage, the coarse-level features will also be aligned and fused again for the final prediction.}
\label{fig:loc}
\end{figure*}

\subsection *{Fovea Localization}
The ranking of fovea localization task is shown in Table \ref{tab:fovea_result}. The results are evaluated by the fovea localization score (see Section \ref{challenge_evaluation}) and Euclidean distance (ED). Teams are ranked by the fovea localization score. The methods of the teams are summarized in Table \ref{tab:fovea_method}. Analogously to the glaucoma grading task, we also implemented a baseline for fovea localization task, which is shown in Figure \ref{fig:loc}. The input of the network is the whole fundus image, and the output is a 2D vector indicating the coordinate of the fovea center. The backbone of the network is ResNet50 and is supervised by the combination of Euclidean distance and MSE loss.

\begin{table}[h]
\centering
\caption{Fovea localization ranking in the GAMMA Challenge.}
\begin{tabular}{c|c|cc}
\hline
\multirow{2}{*}{Rank} & \multirow{2}{*}{Team} & \multirow{2}{*}{Score} & \multirow{2}{*}{ED} \\
                      &                       &                        &                     \\ \hline
1                     & DIAGNOS-ETS           & 9.60294                & 0.00413             \\
2                     & IBME                  & 9.58847                & 0.00429             \\
3                     & SmartDSP              & 9.57458                & 0.00444             \\
4                     & MedIPBIT              & 9.53757                & 0.00485             \\
5                     & Voxelcloud            & 9.53443                & 0.00488             \\
6                     & EyeStar               & 9.51465                & 0.0051              \\
7                     & WZMedTech             & 9.45846                & 0.00573             \\
8                     & MedICAL               & 9.34639                & 0.00699             \\
9                     & FATRI\_AI             & 9.33749                & 0.0071              \\
10                    & HZL                   & 9.22303                & 0.00842             \\ \hline
\end{tabular}
\label{tab:fovea_result}
\end{table}

On fovea localization task, the methods of the teams varies a lot. In the top-10 teams, SmartDSP, MedIPBIT, WZMedTech processed the task as a coordinate regression task, just like we did in the baseline method. VoxelCloud and DIAGNOS-ETS processed the task as a binary segmentation task. They generated a circle centered on the fovea location. The circle is then taken as the segmentation target for the binary segmentation task. The center of the segmented result is finally taken as the fovea location. Eyestar, IBME and MedICAL processed the task as a heatmap prediction task. They generated the ground-truth heatmap by Gaussian kernel. This strategy is similar to the binary segmentation, except it is supervised by a soft target,  which is a normal distribution centered on fovea location. In contrast, FATRI-AI processed the task as a detection task. They generated a 160\texttimes160 square centered on the fovea location and used a YOLO (\cite{redmon2016you}) network to detect the region.

Almost half of the teams utilized a coarse-to-fine multi-stage strategy, including SmartDSP, EyeStar, MedIPBIT, WZMedTech and MedICAL. Most of them cropped Region Of Interest (ROI) based on the coarse stage predictions. The cropped region is then refined by the later stage. EyeStar proposed a more sophisticated architecture based on this strategy and named it Two-Stage Self-Adaptive localization Architecture (TSSAA). They first cropped multi-scale ROI based on the coarse predictions. Then they fused both multi-scale ROI and coarse-level features using sequential ROI Align layer, concatenation, self-attention modules (\cite{vaswani2017attention}) and Fuse layer. An illustration of TSSAA is shown in Figure \ref{fig:loc}.

\subsection *{OD/OC Segmentation}
The ranking of OD/OC segmentation task is shown in Table \ref{tab:segmentation_result}. The results are evaluated by the two Dice values, vertical optic Cup-to-Disc Ratio (vCDR) and the OD/OC segmentation score (see Section \ref{challenge_evaluation}) in the GAMMA Challenge. Teams are ranked by the OD/OC segmentation score. The methods of the teams are summarized in Table \ref{tab:segmentation_result}. A standard UNet is also adopted as the baseline of the task.

\begin{table}[t]
\centering
\caption{OD/OC segmentation ranking in the GAMMA Challenge.}
\resizebox{0.9\textwidth}{!}{%
\begin{tabular}{c|c|cccc}
\hline
\multirow{2}{*}{Rank} & \multirow{2}{*}{Team} & \multirow{2}{*}{Score} & \multirow{2}{*}{Dice-disc(\%)} & \multirow{2}{*}{Dice-cup(\%)} & \multirow{2}{*}{vCDR} \\
                      &                       &                        &                             &                             &                       \\ \hline
1                     & Voxelcloud            & 8.36384                & 96.25                       & 87.84                       & 0.04292               \\
2                     & DIAGNOS-ETS           & 8.3275                 & 95.96                       & 87.74                       & 0.04411               \\
3                     & WZMedTech             & 8.31621                & 96.11                       & 88.04                       & 0.04538               \\
4                     & HZL                   & 8.30093                & 95.83                       & 88.00                       & 0.04562               \\
5                     & SmartDSP              & 8.28488                & 95.79                       & 88.01                       & 0.04642               \\
6                     & MedICAL               & 8.27264                & 95.75                       & 87.57                       & 0.0464                \\
7                     & IBME                  & 8.2309                 & 95.79                       & 87.66                       & 0.04887               \\
8                     & FATRI\_AI             & 8.18773                & 95.40                       & 86.69                       & 0.04917               \\
9                     & MedIPBIT              & 8.15502                & 95.49                       & 87.67                       & 0.05258               \\
10                    & EyeStar               & 8.07253                & 94.77                       & 85.83                       & 0.05326               \\ \hline
\end{tabular}
}
\label{tab:segmentation_result}
\end{table}

\begin{figure*}[h]
\centering
\includegraphics[width=1\linewidth]{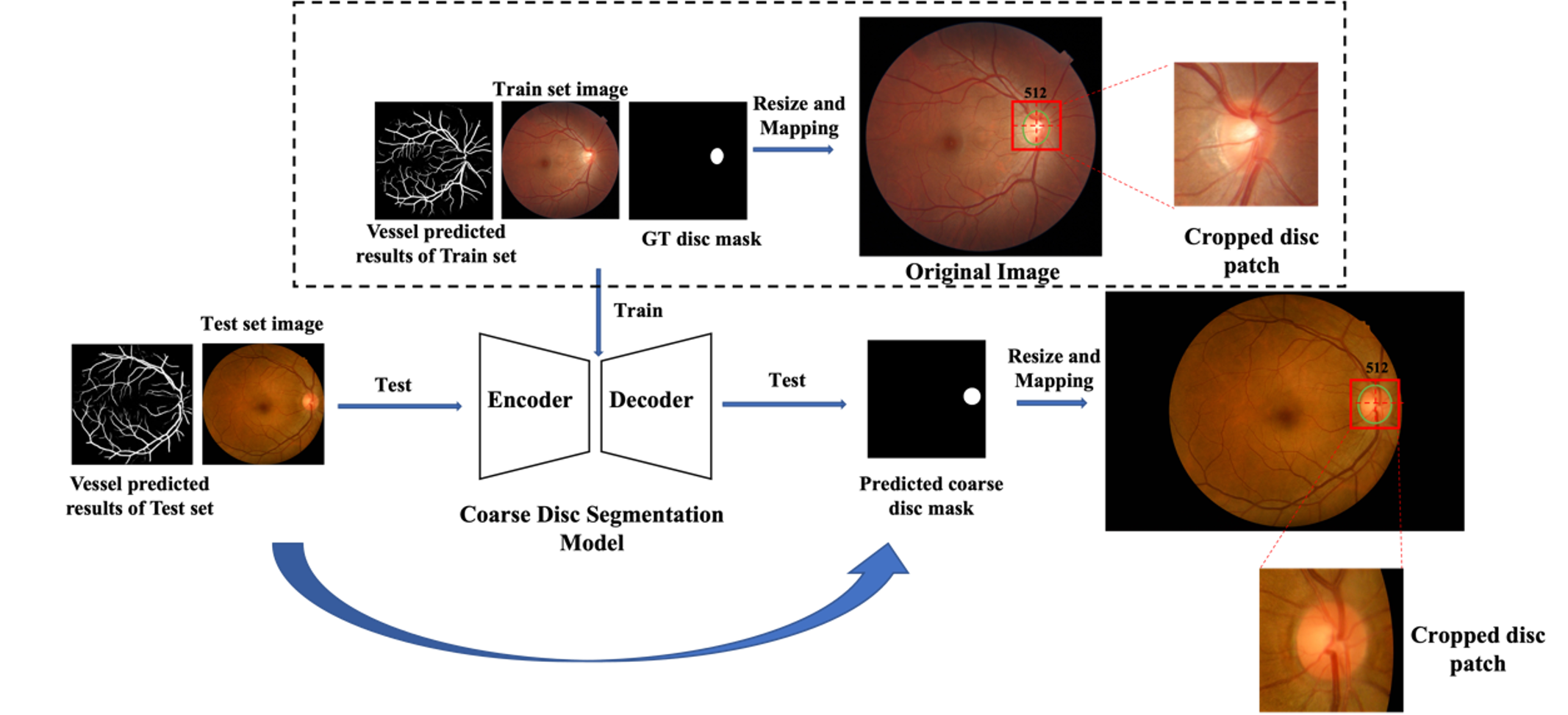}
\caption{The coarse stage of OD/OC segmentation model proposed by VoxelCloud. The blood vessel segmentation results predicted from a pre-trained network will be concatenated as the input for the coarse OD segmentation}
\label{fig:seg}
\end{figure*}

Like the fovea localization task, all the teams except HZL adopted a coarse-to-fine multi-stage strategy. Generally speaking, OD ROI will be first obtained through the coarse OD segmentation stage. The cropped OD patches will be sent to a subsequent Fine-grained OD/OC segmentation network to obtain the final result. Different from the others, VoxelCloud utilized the blood vessel information to improve the OC/OD segmentation. They first used a pre-trained model to obtain the fundus images' blood vessel segmentation masks. The vessel masks are then concatenated with fundus images as the input. An illustration of their method is shown in Figure \ref{fig:seg}.

\begin{table*}[!htbp]
\centering
\caption{Summary of the fovea localization methods in the GAMMA Challenge}
\resizebox{\textwidth}{!}{%
\begin{tabular}{c|c|c|c|c}
\cline{1-5} 
Team        & Architecture  & Preprocessing   & Ensemble   & Method                                                                                                                                                                                                                                                                                                                                        \\ \cline{1-5}
SmartDSP \cite{he2022joined}    & Efficientnet-b4                                                   &  \makecell{(i) Center crop to 2000\texttimes 2000, \\ padding when height \\or width less than 2000\\ (ii) Resize to 224\texttimes 224\\ (iii) Default Data Augmentation}  & 2-fold esmble by averaging          &  \makecell{Two stages coordinate regression:\\  (i) Coarse localization, \\crop to 512\texttimes 512, \\ (ii) Fine-grained localization.}                                                                                                                                                                                      \\  \cline{1-5} 
VoxelCloud  & TransUNet-like architecture                                       &  \makecell{(i) Remove black background\\ (ii) Pad and resize to 512\texttimes 512\\ (iii) Default Data Augmentation + \\Blur + JPEG compression +\\ GaussNoise + Coarse Dropout}                                     & \makecell{Ensemble the predictions \\of 30 models train with \\different hyper-parameters} & \makecell{Binary segmentation the \\fovea centered circle. \\Using the sum of \\binary cross-entropy loss, \\ SoftDice loss, SSIM loss,\\ IOU loss and L1 loss \\to supervise}                                  \\  \cline{1-5} 

EyeStar     & \makecell{Poposed Two-Stage \\Self-Adaptive\\ localization Architecture\\ (TSSAA)} & \makecell{(i) Resize to 998\texttimes998\\(ii) Crop to 896\texttimes896\\(iii) Resize to 224\texttimes224\\(iiii) Default data augmentation}                                                                                                                                                                                                                                 &                                      &  \makecell{Two stages heatmap prediction:  \\(i) Coarse heatmap prediction, \\crop to multi-scale ROI, \\ (ii) Fine-grained localization fusing \\multi-scale ROI and\\ coarse-level features }                                                                                                                                                                                      \\  \cline{1-5} 
HZL         & \makecell{UNet with \\ EfficientNet Backbone}                                    & \makecell{Fundus: Resize to 1024\texttimes1024\\OCT: Resize to 1024\texttimes1024\\ Default  Data Augmentation} &\makecell{Pick 5 best models on 5 \\ different validation folds.\\ Ensemble the results \\ by taking the average.}                                                                                                                                                                                 & \makecell{A multi-task UNet\\ to jointly learn glaucoma grading, \\ OD/OC segmentation \\ and fovea localization. \\Recurrently run the \\model for coarse-to-fine\\ localization}
\\  \cline{1-5} 

MedIPBIT    &  \makecell{ResNet50 for coarse \\localization\\ ResNet101 for Fine-grained\\ localization}  & Resize to 512\texttimes 512  &                                      & \makecell{Three stages coordinate regression:\\ (i) Coarse localization, crop to ROI \\(ii) Sequential two-stage \\Fine-grained localization.} \\  \cline{1-5} 
IBME        &  \makecell{UNet with EfficientNetB5 \\backbone}  &  \makecell{(i) Padding to 2000\texttimes2992 \\(ii) Default Data Augmentation}    &        & \makecell{End-to-end heatmap prediction \\with maximization likelihood \\for the localization} \\  \cline{1-5} 
WZMedTech   &    \makecell{HDRNet (\cite{xie2020end})\\ for the first \\and second stage\\ ResNet50 for the third stage} & \makecell{(i) Center crop to 1920\texttimes1920 \\(ii) Resize to 224\texttimes224 \\(iii) Default Data Augmentation}   &                                      & \makecell{Three stages coordinate regression, \\predicted ROI of last \\stage is cropped as\\ the input of the next stage} \\  \cline{1-5} 
DIAGNOS-ETS &  Double stacked W-Net   &    \makecell{ (i) Resize to 512\texttimes512\\ (ii) Default data augmentation + \\Color Normalization}  & 4-fold temperature ensemble    & \makecell{End-to-end binary segmentation \\the fovea centered circle} \\  \cline{1-5} 
MedICAL     & \makecell{ResNet50 for coordinate \\regression branch\\ EfficientNet-B0 for heatmap \\predication branch}  &  \makecell{(i) Pick G channel of RGB image \\ (ii) Histogram equalization \\(iii) Default data augmentation}  & \makecell{Ensemble the results of \\heatmap branch and\\ coordinate regression branch. \\ If Euclidean distance of \\them larger than 30,\\ take the regression result. \\If else, take the average of \\two results}    & \makecell{Two stages: (i)  Coarse OD/Macular \\segmentation, crop ROI to \\128\texttimes128 and 256\texttimes256 \\ (ii) Feed 128\texttimes128 patches and\\ 256\texttimes256 patches\\to a heatmap predication \\network and coordinate regression\\ network respectively,\\ fuse the results of two branches \\for the final predication} \\ \cline{1-5} 
FATRI-AI    & YOLOv5s (\cite{redmon2016you})       & \makecell{(i) Crop black background \\ (ii) Default data augmentation \\+ Mosaic (\cite{chen2020dynamic})\\ + Cutout} &                                      & \makecell{End-to-end macular \\region detection,\\ macular region is generated by \\a 160\texttimes 160 square \\centered on fovea location}  \\  \cline{1-5} 
\end{tabular}%
}
\label{tab:fovea_method}
\end{table*}

\begin{table*}[!htbp]
\centering
\caption{Summary of the OD/OC segmentation methods in the GAMMA Challenge}
\resizebox{\textwidth}{!}{%
\begin{tabular}{c|c|c|c|c}
\hline
Team        & Architecture                                                                                                                                                          & Preprocessing                                                                                                                                                                    & Ensemble                                                                                                                                                                                                                  & Method                                                                                                                                                                                                                                                                                                \\ \hline
SmartDSP (\cite{he2022joined})    &  \makecell{DeepLabv3 with ResNet34 \\encoder for coarse \\segmentation DeepLabv3 with \\EfficientNet-b2 encoder for\\ Fine-grained segmentation}  &  \makecell{(i) Crop to 512\texttimes 512 centered \\on the highest brightness point\\ (ii) Default Data Augmentation}                                           & 2-fold ensemble by averaging                                                                                                                                                                                             &  \makecell{Two stages:  (i) Coarse OD \\segmentation, cropping\\  (ii) Fine-grained OD/OC segmentation}                                                                                                                                                                        \\ \hline
VoxelCloud  &  \makecell{TransUNet-like architecture \\for coarse segmentation\\  CENet, TransUNet and Segtran \\for Fine-grained segmentation}                &  \makecell{(i) Resize to 512\texttimes 512\\ (ii) Default data augmentation}                                                                                    &  \makecell{5-fold ensemble by \\averaging for coarse segmentation\\  Ensemble the predictions of \\five folds, three networks and \\two kinds of input by\\ averaging for Fine-grained \\segmentation}    &  \makecell{Two stages: (i) Coarse OD \\segmentation taking blood vessel\\ mask concatenated fundus \\image as input, cropping\\  (ii) Fine-grained OD/OC segmentation \\taking cropped patches \\and polar transformed \\patches as inputs.\\ Model supervised by \\BCE loss + Dice loss}  \\ \hline

EyeStar     & \makecell{Segtran (\cite{li2021medical}) with \\EfficientNet-B4 backbone}                                                                                                                                 &  \makecell{(i) Crop to 576\texttimes576 disc \\region by \\MNet DeepCDR (\cite{fu2018joint}) \\ (ii) Resize to 288\texttimes288\\ (iii) Default data augmentation}  &                                                                                                                                                                                                                          & \makecell{Two stages: (i) Coarse OD \\segmentation using \\ CNN, cropping\\  (ii) Fine-grained OD/OC segmentation \\using Segtran}                                                                                                                                                                                                                                                                                                    \\ \hline
HZL         & \makecell{UNet with \\ EfficientNet Backbone}                                    & \makecell{Fundus: Resize to 1024\texttimes1024\\OCT: Resize to 1024\texttimes1024\\ Default  Data Augmentation} &\makecell{Pick 5 best models on 5 \\ different validation folds.\\ Ensemble the results \\ by taking the average.}                                                                                                                                                                                 & \makecell{A multi-task UNet\\ to jointly learn glaucoma grading, \\ OD/OC segmentation \\ and fovea localization.\\ FAM (\cite{huang2021fapn}) is \\adopted for \\the better segmentation } \\ \hline
MedIPBIT    &  \makecell{CNN-Transformer Mixed UNet\\ CNN backbone implemented \\by ResNet34}                                                              & Resize to 512\texttimes512                                                                                                                                                                &                                                                                                                                                                                                                          &  \makecell{Two stages:  (i) Coarse OD\\ segmentation, cropping\\  (ii) Fine-grained OC segmentation}                                                                                                                                                                           \\ \hline
IBME        &  \makecell{UNet with \\EfficientNetB3 backbone \\for OC center localization\\ UNet with \\EfficientNetB6 backbone \\for Fine-grained segmentation}   & Default data augmentation                                                                                                                                                        &                                                                                                                                                                                                                          &  \makecell{Two stages:  (i) OC center \\localization, crop ROI \\to 512\texttimes 512\\  (ii) Fine-grained OD/OC segmentation}                                                                                                                                                           \\ \hline
WZMedTech   &  \makecell{DeepLabV3 for\\ coarse segmentation\\ TransUNet for\\ Fine-grained segmentation}     &  \makecell{(i) Center crop to 1920\texttimes 1920\\ (ii) Default data augmentation }                                                                          &  \makecell{In the Fine-grained stage, \\ensemble the models supervised\\ by cross-entropy loss  + \\boundary loss + dice loss\\ and that supervided by focal loss + \\dice loss by taking the average}  &  \makecell{Two stages:  (i) Coarse OD \\segmentation, crop ROI \\to 512\texttimes 512\\  (ii) Fine-grained OD/OC segmentation}                                                                                                                                                           \\ \hline
DIAGNOS-ETS & Double stacked W-Net                                                                                                                                                  &  \makecell{(i) Resize to 512\texttimes512\\ (ii) Default data augmentation + \\Color normalization}                                                              &  \makecell{In coarse OD segmentation: \\4-fold ensemble by taking average\\ In Fine-grained OD/OC segmentation: \\4-fold temperature ensemble}                                                      &  \makecell{Two stages:  (i) Coarse OD \\segmentation, crop ROI \\to 512\texttimes 512\\  (ii) Fine-grained OD/OC segmentation}                                                                                                                                                           \\ \hline
MedICAL     & \makecell{UNet with EfficientNet-B4 \\backbone}                                                                                                                                    &  \makecell{(i) Resize to 512\texttimes512\\ (ii) Default data augmentation}    &                                    &  \makecell{Three stages: (i) Coarse OD/Macular\\ segmentation, crop OD ROI \\to 448\texttimes448\\ (ii) Fine-grained OD/OC segmentation, \\crop OC ROI TO 256\texttimes256\\ (iii) Fine-grained OC segmentation}  \\ \hline
FATRI-AI    &  \makecell{YOLOv5s for coarse segmentation\\ HRNet for Fine-grained segmentation}                                                            &  \makecell{(i) Resize to 608\texttimes608\\  (ii) Default data augmentation \\+ Mosaic (\cite{chen2020dynamic})\\ + Cutout}                                                           &                                                                                                                                                                                                                          &  \makecell{Two stages: (i) Coarse OD \\segmentation, crop ROI \\to 512\texttimes 512\\  (ii) Fine-grained OD/OC segmentation. \\Final results will be \\smoothed as ellipses}                                                                                                               \\ \hline
\end{tabular}%
}
\label{tab:segmentation_method}
\end{table*}

\end{document}